\documentclass[review]{elsarticle}
\usepackage[whole]{bxcjkjatype}
\usepackage{bm}
\usepackage{amssymb}
\usepackage{amsmath}
\usepackage{color}

\usepackage[ruled,vlined]{algorithm2e}
\usepackage{algorithm2e,setspace}

\SetCommentSty{mycommfont}
\usepackage{booktabs}
\usepackage{graphicx}
\usepackage{floatrow}
\usepackage{array}
\newcolumntype{C}[1]{>{\centering\arraybackslash}p{#1}}
\usepackage{hyperref}
\usepackage{subfigure}
\newcommand{\tabref}[1]{Table~\ref{#1}}
\newcommand{\equref}[1]{Eq.~(\ref{#1})}
\newcommand{\figref}[1]{Fig.~\ref{#1}}

\newcommand{\algref}[1]{Algorithm~\ref{#1}}

\DeclareMathOperator*{\argmax}{argmax}

\makeatletter
\def\ps@pprintTitle{%
  \let\@oddhead\@empty
  \let\@evenhead\@empty
  \def\@oddfoot{\reset@font\hfil\thepage\hfil}
  \let\@evenfoot\@oddfoot
}
\makeatother

\begin{document}

\begin{frontmatter}

\title{
    Cyclic Policy Distillation: Sample-Efficient Sim-to-Real Reinforcement Learning with Domain Randomization
}

\author[naistaddress]{Yuki Kadokawa\corref{cor1}}
\ead{kadokawa.yuki.kv3@is.naist.jp}

\author[naistaddress]{Lingwei Zhu} \ead{zhu.lingwei.zj5@is.naist.jp}

\author[naistaddress]{Yoshihisa Tsurumine} \ead{tsurumine.yoshihisa@is.naist.jp}

\author[naistaddress]{Takamitsu Matsubara}
\ead{takam-m@is.naist.jp}

\cortext[cor1]{Corresponding author}

\address[naistaddress]{Graduate School of Science and Technology, Division of Information Science, Nara Institute of Science and Technology, 8916-5 Takayamacho, Ikoma, Nara, Japan}

\begin{abstract}
Deep reinforcement learning with domain randomization learns a control policy in various simulations with randomized physical and sensor model parameters to become transferable to the real world in a zero-shot setting. 
However, a huge number of samples are often required to learn an effective policy when the range of randomized parameters is extensive due to the instability of policy updates. 
To alleviate this problem, we propose a sample-efficient method named cyclic policy distillation (CPD). 
CPD divides the range of randomized parameters into several small sub-domains and assigns a local policy to each one. 
Then local policies are learned while cyclically transitioning to sub-domains. CPD accelerates learning through knowledge transfer based on expected performance improvements. 
Finally, all of the learned local policies are distilled into a global policy for sim-to-real transfers. 
CPD's effectiveness and sample efficiency are demonstrated through simulations with four tasks (Pendulum from OpenAIGym and Pusher, Swimmer, and HalfCheetah from Mujoco), and a real-robot, ball-dispersal task.
We published code and videos from our experiments at \url{https://github.com/yuki-kadokawa/cyclic-policy-distillation}.
\end{abstract}

\begin{keyword}
\texttt domain randomization \sep sim-to-real \sep deep reinforcement learning
\end{keyword}

\end{frontmatter}

\section{Introduction}
    Deep reinforcement learning (DRL) is one of the most promising methods for robots to automatically acquire autonomous control policies through interaction with a real-world environment \cite{dqn,google-picking,drl-door}.
    Recent studies utilize DRL to learn control policies in simulations and thus make these policies transferable to the real world, an approach called sim-to-real.

    In sim-to-real by DRL, zero-shot learning \cite{DR-origin,DR-deformable,ActiveDR}, in which policies are learned in a simulation environment and utilized in real-world environments, has been the focus of attention. However, in zero-shot learning, policies must be learned with domain randomization (DR) in which the simulation parameters are widely randomized to reduce the reality gap between the real-world and simulation environments. As required tasks have become increasingly complex in recent years, the randomization range and the number of parameters to be addressed by DR have also increased. The learning policies in such large domains have been difficult due to gradient instability and the increased number of learning samples that are required \cite{P2PDRL}. 
    
    Therefore, few-shot learning \cite{NeuralSim,sim2real-si,sim2real2sim}, such as differentiable simulators and real-to-sim-to-real, have been proposed where the simulation parameters are adjusted to reduce the use of simulators by utilizing a certain amount of real-world samples. These approaches can learn policies with fewer samples than DRL with DR, assuming that the learning setup can be constructed in a real-world environment. On the other hand, since these approaches require interaction with real-world environments, they eventually become human-in-the-loop settings, or applying them to such tasks as sand and chemical powder tasks will be complicated because real environments are difficult to set up or initialize.

    The above background suggests the significance of re-focusing and further developing DRL with DR to improve its sample efficiency and learning stability and achieving zero-shot learning that is applicable to more complex tasks. Establishing such development is challenging in real environments.

    This study addresses this zero-shot approach.
    In particular, such use of RL with DR is classified into visual randomization \cite{DR-origin,visual-DR} and dynamics randomization. We tackled the latter \cite{P2PDRL,DiDoR}. 

    Sample efficiency, evaluated by the number of samples required to reach a certain performance threshold, is one of the biggest DRL subjects. In the category of DRL with DR, where full-domain denotes a complete range of randomized parameters and sub-domain denotes part of a full-domain, the existing methods for improving sample efficiency fall into three categories: limiting the range of randomized parameters \cite{simopt-origin,simopt-baysian,bayessim}; making a curriculum that extends the domain range to a full-range \cite{ADR}; dividing the full-domain into several sub-domains and learning local policies restricted to each sub-domain \cite{DiDoR,P2PDRL,DnC}. These categories suffer from such issues as a lack of real-world samples for a posterior that adjusts the ranges, makes a curriculum based on domain difficulty, or creates inefficiency resulting from independence among sub-domains.

    Consequently, if there were a mechanism that shared samples or learned values and policies among sub-domains, it might be possible to improve sample efficiency. 
    Nevertheless, naively sharing may even degrade the performance due to domain gaps among sub-domains. Therefore, we are motivated to develop a new scheme that appropriately shares learned values and policies among sub-domains to improve both sample efficiency and learning performance.
    If sub-domains share similar parameters, the resulting Markov decision processes (MDPs) might also resemble. 
    Under this assumption, the values and policies in similar sub-domains probably share similar characteristics. 
    Therefore, we hypothesize that learning can be accelerated by transferring among neighboring partitioned sub-domain's values and policies. 
    This assumption will be validated empirically through our experiments.

\begin{figure}
    \vspace{1.6mm}
    \hspace{37mm}
    \centering
    \includegraphics[width=0.85\columnwidth]{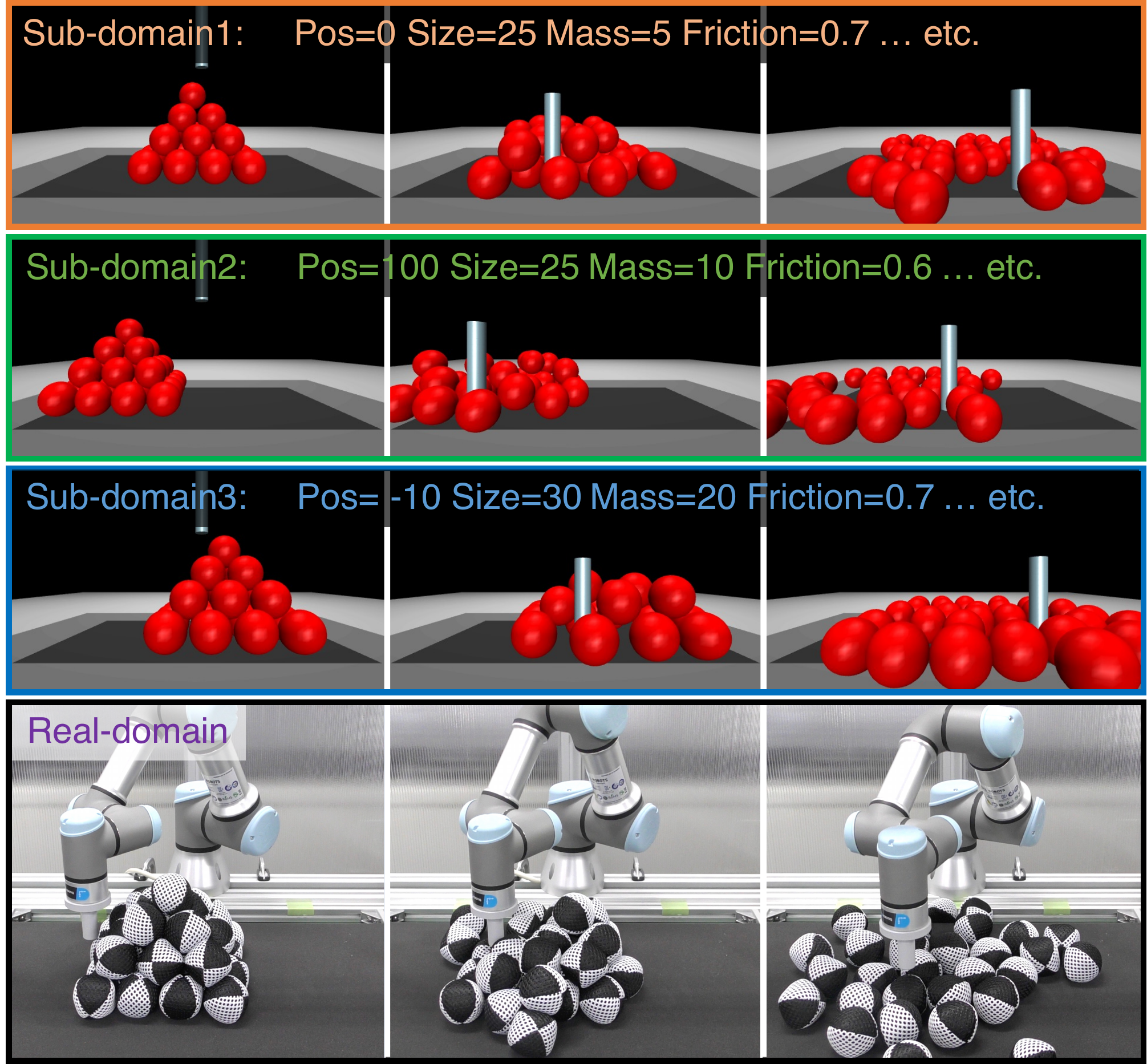}
    \caption{
        Experimental scenes of robotic ball-dispersal task with domain randomization: Upper three rows are simulation environments; bottom row is real-world environment.
        Simulation parameters are randomized, including ball-size (radius) and ball-position (center position of all balls).
        Sub-domains are made from a simulation environment formed by dividing the range of randomized parameters. 
        Learned policies in all sub-domains are distilled into a global policy transferred to a real domain for sim-to-real transfer.
    }
    \label{fig:senzai}
\end{figure}

    This paper proposes cyclic policy distillation (CPD) as a sample-efficient RL method for sim-to-real in a zero-shot setting. CPD divides the range of randomized parameters into several small sub-domains and assigns a local policy to each one.
    Then the learning of local policies is performed while {\it cyclically} transitioning the target sub-domain to neighboring sub-domains, and the learned values and policies of the neighboring sub-domains are exploited with a monotonic policy-improvement scheme.
    Finally, all of the learned local policies are distilled into a global policy for a sim-to-real transfer. To validate CPD's effectiveness and sample efficiency, we evaluated it on simulation problems, i.e., with four tasks (Pendulum from OpenAIGym \cite{gym} and Pusher, Swimmer, and HalfCheetah from Mujoco \cite{mujoco}), and a real-robot ball-dispersal task (\figref{fig:senzai}). 
    Ball-dispersal tasks are designed as a simplified environment for operations that handle multiple particulate objects, such as powder manipulation \cite{powder-sim2real, powder-manipulation, powder-manipulation-learning} and soil excavation \cite{excavator-sim2real,excavator-sim2real-RL,excavator-sim2real-geometric}. Simplification is achieved using increased ball-size with a small number of balls; however, difficulties remain, including applying DRL due to the hardness of the automatic initialization and the enormous patterns of observation samples with their irregular shapes and deformations. Therefore, it is a suitable task to be tackled by a sim-to-real approach.
    
    The rest of this paper is organized as follows.
    Section \ref{related-works} describes related works. 
    In Section \ref{preliminaries}, we offer preliminary considerations before discussing our proposed CPD method, which
    Section \ref{proposed-method} describes. 
    Section \ref{simulation-experimens} presents simulation experiments, and 
    Section \ref{real-experiments} presents real-robot experiments. 
    In Section \ref{discussion}, we discuss the limitations of the CPD method.
    Finally, Section \ref{conclusion} concludes this paper.

\begin{figure}[t]
    \vspace{-1.2mm}
    \centering
    \includegraphics[width=0.99\columnwidth]{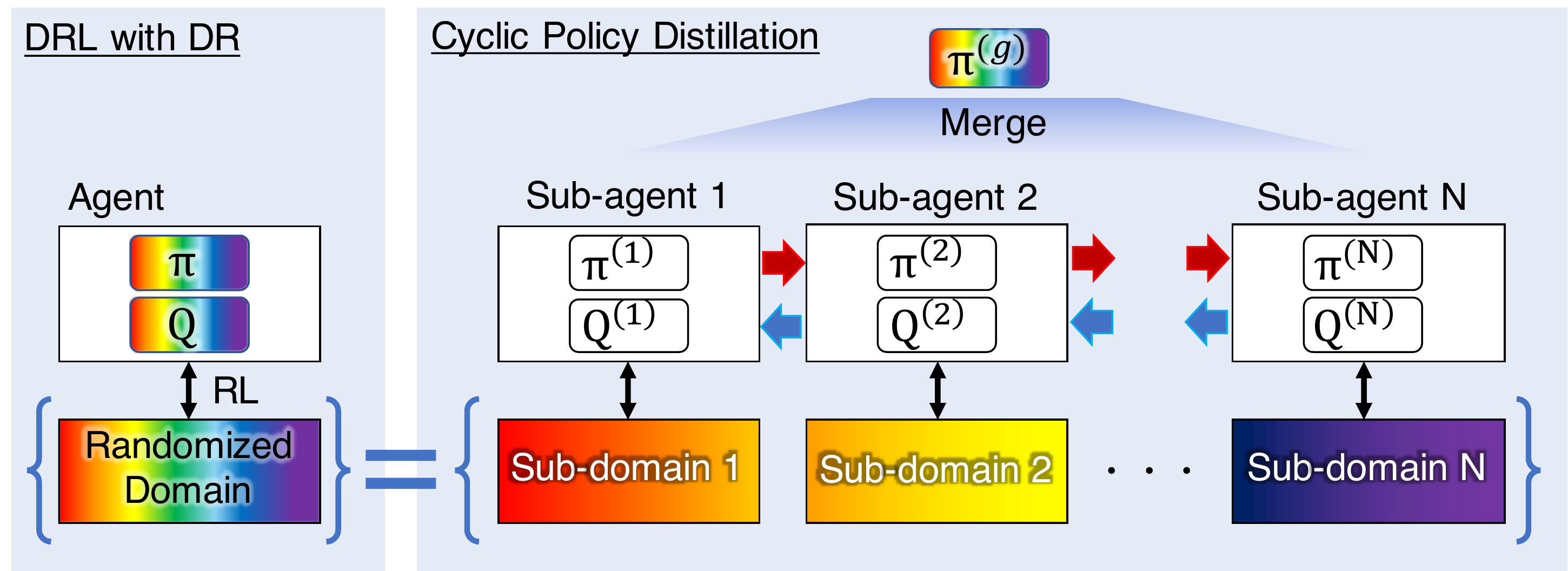}
    \caption{
    Overview of cyclic policy distillation: Unlike typical methods that conduct DR in simulations with a full range of randomized parameters, CPD divides the range of randomizing parameters into several small sub-domains and assigns local policy $\pi^{(n)}$ and local value $Q^{(n)}$ to each sub-domain $n$.
    Next the learning of local policies is performed while {\it cyclically} transitioning the target sub-domain to neighboring sub-domains and exploiting the learned values and policies of neighboring sub-domains with a monotonic policy-improvement scheme. 
    Finally, all learned local policies are distilled into one global policy $\pi^{(g)}$ for sim-to-real transfer.
    }
    \label{fig:proposed-overview}
\end{figure}

\section{Related Works}
    \label{related-works}
    Reinforcement learning (RL) in sim-to-real can be divided into two categories, depending on whether the real-world environment is utilized for learning policies: few-shot learning and zero-shot learning.
    
    \subsection{Few-shot sim-to-real}
        In the few-shot settings of RL, recent studied have proposed these two approaches.
        
        \textbf{Differentiable simulator (DS):}
            In this method, the simulator behavior is directly updated by back propagation by real-world samples using differentiable simulators \cite{NeuralSim,DiffTaichi}.
            Since the methods reproduce the simulator's behavior using differentiable physical equations, non-differentiable physics has to be approximated with other models, such as NNs, or learned with robust policies.

        \textbf{System identification (SI):}
            This approach performs test behaviors in a real-world environment using the policies obtained in simulations \cite{bayessim,sim2real-si,sim2real2sim}. The range of randomized parameters is shifted and limited by a probabilistic distribution to cover the plausible parameters based on the errors between state-action sequences in simulations and in the real world. Since executing actions in the real-world is necessary, such a scope is beyond the scope of this paper, which aims for a zero-shot setup.

    \subsection{Zero-shot sim-to-real}
        \label{stable-DR}
        In zero-shot settings of RL with DR, some studies have attempted to stabilize the learning process.
        When DR frequently switches the randomized parameters, there is considerable variation in the samples, and the updating direction of the policies can also frequently change. Thus, the learning of the policies becomes unstable, resulting in low sample efficiency and poor performance \cite{P2PDRL}.
        Previous works have stabilized the updating policies by dividing the range of randomized parameters into smaller ranges to avoid frequent changes in the policies. 
        The following four major approaches have been pursued for this purpose. 

\begin{table}[t]
    \begin{center}
        \caption{
           Characteristics of proposed method and related works: 
           Using real samples denotes methods using samples from real-world environments.
           Parameter tuning denotes methods that tune simulation environment's parameters.
           Design curriculum denotes methods that update the range of DR with a designed curriculum.
           Dist./Partition denotes methods using policy distillation and partitioning parameters, Know.-trans. denotes methods learning local policies/values with transferring knowledge between local policies/values, Range-adjust denotes methods optimizing parameter distribution, Scheduling denotes methods learning local policies/values with scheduled transitions of sub-domains.
           \label{table:previous_compare}
        }
        \begin{tabular}{lccccccc}
            \toprule
                \textbf{Method} & \textbf{DS} & \textbf{SI} & \textbf{DiDoR} & \textbf{P2P} & \textbf{DnC} & \textbf{ADR} & \textbf{CPD} \\
            \midrule
                \textbf{Using real samples} & \checkmark & \checkmark & - & - & - & - & - \\
                \textbf{Parameter tuning} & \checkmark & \checkmark  & - & - & - & - & - \\
                \textbf{Design curriculum} & - & -  & - & - & - & \checkmark & - \\
                \textbf{Dist./Partition} & - & - & \checkmark & \checkmark & \checkmark & - & \checkmark \\
                \textbf{Know.-trans.} & - & -   & - & \checkmark & \checkmark & - & \checkmark \\
                \textbf{Scheduling} & - & -  & - & - & - & \checkmark & \checkmark \\
            \bottomrule
        \end{tabular}
    \end{center}
\end{table}

        \textbf{Distilled domain randomization (DiDoR) \cite{DiDoR}:}
            In this approach, the learning of local policies and their generalization to global policy are performed separately.
            First, each local policy is learned independently.
            After the learning of all local policies has converged, the policies are distilled into a single global policy. 
        
        \textbf{Peer-to-peer distillation reinforcement learning (P2PDRL) \cite{P2PDRL}:}
            In this approach, local policies are learned and generalized simultaneously to a single global policy.
            After a certain number of learning iterations are completed with a generalization that distills each local policy, all the local policies change their learning range of randomized parameters. By repeating this process, the local policies gradually become a global policy that works in every range of the randomized parameters.
        
        \textbf{Divide-and-conquer (DnC) \cite{DnC}:}
            This approach alternates between learning local policies and generalizing them into a single global policy.
            After all of the local policies have been learned for a certain number of samples, they are distilled into a global policy for generalization. These procedures are repeated until the learning converges. 

        \textbf{Active domain randomization (ADR) \cite{ADR}:}
            In this approach, the domain range's breadth can be classified as a curriculum, and the domain range is expanded adaptively as the performance of the policy improves.
            A reward threshold must be defined in advance to evaluate whether the policy has achieved satisfactory performance for the domain's current size. Once the reward threshold is reached, the domain ranges are expanded, eventually extending the scope to the full range of domains.
            In general, curriculum design requires deep knowledge of the task, which is expensive to implement \cite{curriculum-all,curriculum-water}.

        A common feature of the above previous methods (except CL) is that the sub-domains to be learned are selected {\it randomly}. Therefore, the gap of the sub-domains in distillation can be large, making the learning results unstable and failing to improve the sample efficiency. In contrast to these methods, our proposed method {\it cyclically} transitions to neighboring sub-domains with similar ranges of randomized parameters and exploits the neighboring sub-domains' learned values and policies with a monotonic policy-improvement scheme. Consequently, values and policies are shared within small-gap domains to ensure improved learning stability and sample efficiency. 
        The characteristics of our proposed method and related works are summarized in \tabref{table:previous_compare}.

        Since a feature between CPD and ADR on behalf of CL (the reward threshold in ADR) is changed by both the task and domain characteristics, designing it is difficult to uniquely determine the threshold. Thus the method cannot guarantee that its policies will learn in the full-domain. On the other hand, our proposed CPD requires some knowledge to determine the target of the domain segmentation, as we will discuss below, although it is guaranteed that the policy will learn the full-domain because sub-domains completely learn it.

\section{Preliminaries}
    \label{preliminaries}
    
    \subsection{Reinforcement Learning with Domain Randomization}
        Reinforcement learning (RL) problems are typically formulated by Markov decision processes (MDP) expressed by components $(\mathcal{S},\mathcal{A}, \mu, \mathcal{P}, r,  \gamma)$.
        $\mathcal{S}$ is the set of observations that can be obtained from the environment, and $\mathcal{A}$ is the set of selectable actions.
        $\mu$ is the initial state distribution.
        $\mathcal{P}^a_{s s'}$ denotes the probability of transitioning to observation $s' \in \mathcal{S}$ when action $a \in  \mathcal{A}$ is chosen, given observation $s \in \mathcal{S}$. 
        The reward for making the transition is represented by $r^a_{s s'}$, and $\gamma \in [0,1)$ is the discount factor.
        Policy $\pi(a|s)$ is the probability of choosing action $a$ given $s$.
        We define the stationary distribution induced by $\pi$ as $d_{\pi,\mu}(s) = (1-\gamma)\sum_{t=0}^{\infty}\gamma^{t}\mathcal{P}({s_{t} = s|\mu, \pi})$.
        
        For each observation $s$, evaluation criterion $V^{\pi}$  under policy $\pi$ can be defined: 
        \begin{eqnarray}
            \label{V_function}
            \begin{aligned}
                V^{\pi}(s)={\mathbb{E}}_{\pi, \mathcal{P}}\bigg[\sum_{\substack{t = 0}}^{\infty} \gamma^{t} r_{s_{t}} \bigg| s_{0}= s \bigg],
            \end{aligned}
        \end{eqnarray}
        where $r_{s_{t}} \!=\! \sum_{\substack{a_t \in \mathcal{A} \\ s_{t+1} \in \mathcal{S}}}\pi(a_t|s_t)\mathcal{P}^{a_t}_{s_t s_{t+1}}r^{a_t}_{s_t s_{t+1}}$.
        The goal of RL is to find optimal policy $\pi^{*}$ that maximizes the discounted total reward starting from arbitrary initial observation $s_0$ drawn from $\mu$: $J^{\pi^{*}}_{\mu} = \sum_{s_0} d_{\pi^{*},\mu}(s_0) V^{*}(s_0).$
        It is a well-known result that such an optimal policy satisfies the Bellman equation:
        \begin{eqnarray}
            \label{V_Bellman}
            \begin{aligned}
                V^{*}(s) = \displaystyle\max_{\pi} \sum_{\substack{a \in \mathcal{A}, s' \in \mathcal{S}}} \pi(a|s) \mathcal{P}_{ss'}^{a} \big(r_{ss'}^{a} + \gamma V^{*}(s')\big), 
            \end{aligned}
        \end{eqnarray}
        where $V^{*}(s)$ is the optimal value function.
        To evaluate policies based not only on observation $s$ but also action $a$, the optimal value function is defined:
        \begin{eqnarray}
            \label{Q_Bellman}
            \begin{aligned}
                Q^{*}(s, a) \hspace{-0.05cm}=\hspace{-0.05cm} \displaystyle\max_{\pi} \hspace{-0.05cm} \sum_{s' \in \mathcal{S}}\mathcal{P}_{ss'}^{a}\big(r_{ss'}^{a} \hspace{-0.05cm}+\hspace{-0.05cm}  \gamma \hspace{-0.1cm}  \sum_{a' \in \mathcal{A}} \hspace{-0.05cm} \pi(a'|s')Q^{*}(s', a')\big). \hspace{-0.25cm}
            \end{aligned}
        \end{eqnarray}
        Given $Q$, we can extract its greedy policy by greedy operator $\mathcal{G}(Q^{\pi}) \!:=\!\argmax_{\pi}Q^{\pi}(s,a)$.
        The relationships of Eqs. (\ref{V_Bellman}) and (\ref{Q_Bellman}) also hold for arbitrary stationary policy $\pi$ by removing the $\max$ operator.
        We also define advantage function $A^{\pi}(s,a) \!=\! Q^{\pi}(s,a) \!-\! V^{\pi}(s)$ as the evaluation of how well an action performs compared to the average. 
        The expected advantage can be defined as the expectation of $A^{\pi}(s,a)$ w.r.t. another policy $\pi'$: $A^{\pi'}_{\pi}(s)= \sum_{a}\pi'(a|s)A_{\pi}(s,a).$
        
        As for RL with DR, the transition probability depends on a domain-specific parameter. Then assume $N$ randomly sampled domains, each of which has its own specific transition probability, as $\mathcal{P}_{ss'}^{a,(n)}$ where $n \in N$. The goal of RL with DR is to find optimal policy $\pi^{*}$ that {\it commonly} maximizes the following equation for all $N$ domains: 
        \begin{equation}
            \label{eq:Q_domain}
            \begin{aligned}
            Q^{*, (n)}(s, a) \hspace{-0.05cm}=\hspace{-0.05cm} \displaystyle\max_{\pi} \hspace{-0.05cm} \sum_{s' \in \mathcal{S}}\mathcal{P}_{ss'}^{a, (n)}\big(r_{ss'}^{a} \hspace{-0.05cm}+\hspace{-0.05cm}  \gamma \hspace{-0.1cm}  \sum_{a' \in \mathcal{A}} \hspace{-0.05cm} \pi(a'|s')Q^{*, (n)}(s', a')\big), \hspace{-0.25cm}
            \end{aligned}
        \end{equation}
        where all the domains share the same reward function. Each domain possesses its own policy and value function. 
    
    \subsection{Monotonic Policy Improvement in RL}
        In the presence of various errors and noises, such as approximation error and observation noise, there is no guarantee that the updated RL policy can improve performance. 
        Degraded performance threatens such physical systems as robots.
        
        Drawing inspiration from the classic but popular monotonic improvement (MI) literature \cite{CPI,LPI,SPI}, the problem of performance degradation can be solved if one can ensure that updated policy $\pi_{k+1}$ attains non-negative improvement $\Delta_{k+1}$ against $\pi_k$, i.e., $J^{\pi_{k+1}}_{\mu} - J^{\pi_{k}}_{\mu} \geq \Delta_{k+1} \geq 0$.
        Conservative policy iteration (CPI) \cite{CPI} is one such method that guarantees lower-bounded improvement $\Delta_{k+1} \!\geq\! 0$ by linearly interpolating both the current and greedy policies:
        \begin{align}
            \label{eq:CPI-update}
                    \pi_{k+1} \gets (1-m_{k+1}) \pi_{k} + m_{k+1} \mathcal{G}(Q^{\pi}),
        \end{align}
        where $m_{k+1}\!\in\![0,1]$ is the policy mixture rate.
       Since policy-improvement lower bound $\Delta_{k+1}$ is a negative quadratic function in $m_{k+1}$ \cite{CPI}, we can solve for maximizer $m^*_{k+1}$:
        \begin{align}
            \label{eq:CPI-update-dicide-optimal-rate}
                    m_{k+1}^* = \displaystyle \frac{1-\gamma}{4 R} \sum_{s \in \mathcal{S}} d_{\pi_{k},\mu}(s) \sum_{a \in \mathcal{A}} \pi_{k+1}(a|s) A^{\pi_k}(s,a),
        \end{align}
        where $R$ is the maximum possible reward.
        
    \subsection{Policy Distillation}
        \label{PD}
        Policy distillation (PD) \cite{policy-distillation} for integrating multiple policies into a single policy is used in the proposed CPD method and in previous works described in Section \ref{stable-DR}. Modern deep learning features large-scale models that contain millions of parameters. The network size with the parameters can be compressed in significant portions without performance degradation by PD \cite{policy-distillation}.

        Hence, it is advantageous to compress large-scale models into smaller ones to save computational resources without degrading performance.
        Motivated by the need for compression, the core function of PD is to express the knowledge of a large policy network by a much smaller one \cite{policy-distillation,PD-compress-evaluate}.
        Besides a compressive purpose, PD has natural applications in multi-task RLs, where the policies learned in each sub-task are merged into a single policy to achieve comparable performance on all the sub-tasks \cite{Distral,DistAfterLearn,mult-task-PD}.
        In this paper, our proposed method performs these two distillation tasks to transfer learned knowledge between local policies in divided sub-domains, followed by the final integration of all the local policies into a single global policy. 

\section{Cyclic Policy Distillation}
\label{proposed-method}
    To alleviate the sample inefficiency of RL with DR, we propose a sample-efficient method named cyclic policy distillation (CPD) whose overview is shown in \figref{fig:proposed-overview}. 
    CPD consists of the following three steps:
    \begin{enumerate}
        \item Dividing the randomized domain into sub-domains. 
        \item Cyclic learning of all the local policies.
        \item Distilling local policies into a single global policy.
    \end{enumerate}
    Details of these steps are described below, and the pseudo-code is provided in \algref{algorithm}.
    The CPD parameters are shown in \tabref{table:parameter_setting}.

    \subsection{Dividing a Randomized Domain into Sub-domains}
        \label{dividing} 
\begin{figure}[t]
    \centering
    \includegraphics[width=0.85\columnwidth]{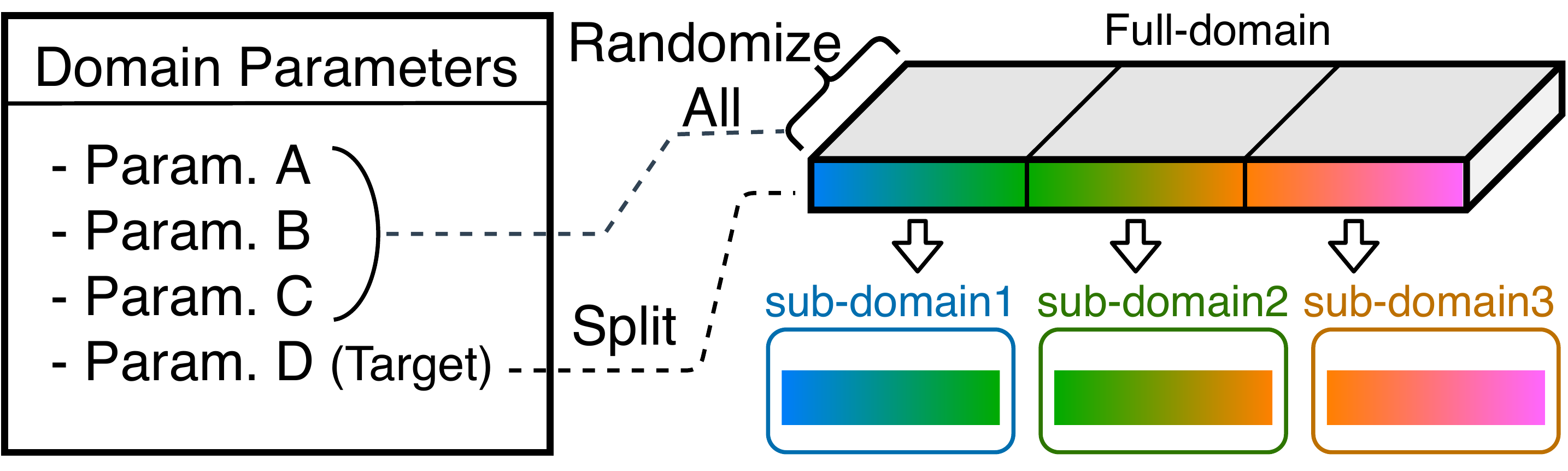}
    \caption{
        Splitting domain parameters into sub-domains:
        In this figure, all parameters are 4 (Parameters A to D).
        Target split parameter is Parameter D.
        Sub-domains are defined as a split range of Parameter D with others randomized in all-ranges.
        As a supplement, number of splitting parameters and splitting methods are arbitrary. Experiment investigates the splitting effect.
    }
    \label{fig:how-to-divide-domain}
\end{figure}

        As shown in \figref{fig:how-to-divide-domain}, CPD divides a full-domain with all the randomized parameters into sub-domains with a part of the randomized parameters based on the target domain parameter and assigns a local policy to each sub-domain. To this end, we set target parameters and their randomization ranges and divide them into $N$ parts, each of which is utilized as the randomization range of each sub-domain. The sub-domains with adjacent index numbers $n \in \{ 1,...,N\}$ should be set up so that their randomization ranges are similar; this is a key arrangement for subsequent steps of CPD, as explained below. 
        
        Although the number of divisions $N$ can be arbitrarily defined, there may be a trade-off between sample efficiency and stability since a small $N$ increases the range of each sub-domain, causing unstable learning, and vice versa. 

\begin{algorithm}[t]
    \SetKwData{Left}{left}\SetKwData{This}{this}\SetKwData{Up}{up}
    \SetKwFunction{Union}{Union}\SetKwFunction{FindCompress}{FindCompress}
    \SetKwInOut{Input}{input}\SetKwInOut{Output}{output}
    \caption{Cyclic Policy Distillation}         
    \SetKwFunction{UpPN}{PolicyUpdate}
    \SetKwFunction{DC}{DataCollect}
    \SetKwProg{Fn}{Function}{:}{\KwRet}
    \label{algorithm}
    \# Set parameters described in \tabref{table:parameter_setting} \\
    \# Separate the domain into sub-domains into $N$ \\
    \# Set networks of policies and value functions \\
    \# Set replay buffers $\mathcal{D}$ of sub-domains \\
    
    \While{until converge}
    {
        \For{$ n = 1, 2, ..., N,\ \ N,N-1,..., 1$}
            {
            \For{$ e = 1, 2, ..., E$}
            {
                \For{$ t = 1, 2, ..., T$}
                {
                    \# Get observation $s_{t}^{(n)}$, reward $r^{(n)}$ \\
                    \# Take action $a_t^{(n)}$ with policy $\pi^{(n)}$ \\
                    \# Push $(s_t^{(n)}, a_t^{(n)},r^{(n)})$ to $\mathcal{D}^{(n)}$ \\
                    \# Update value function $Q^{(n)}$ by RL \\
                    \# Update policy $\pi^{(n)}$ by \equref{eq:proposed-local-loss} \\
                }
            }
            \# Copy network parameters of $Q^{(n)}$ to next sub-domain's $Q$ \\
        }
    }
    \While{until converge}
    {
        \For{$ n = 1, 2, ..., N$}
        {
            \For{$ t = 1, 2, ..., T$}
            {
                \# Get observation $s_{g}^{(n)}$ \\
                \# Take action $a_{g}^{(n)}$ with policy $\pi^{(g)}$ \\
                \# Push $(s_{g}^{(n)}, a_{g}^{(n)})$ to $\mathcal{D}^{(g)}$ \\
                \# Distill policy $\pi^{(g)}$ with $\pi^{(n)}$ by \equref{eq:proposed-Dist-marge} \\
            }
        }
    }
\end{algorithm}

    \subsection{Cyclic Learning of All Local Policies}
        \label{learning}
        For each sub-domain, we define its sub-agent, including local policy $\pi^{(n)}$ and local value function $Q^{(n)}$, where $n$ is an index of the sub-domain. 
        Next the learning of local policies is performed while {\it cyclically} transitioning the target sub-domain to neighboring sub-domains, and the learned values and policies of the neighboring sub-domains are exploited with a monotonic policy-improvement scheme \cite{CPI,LPI,SPI}. 
        The details of the learning process's flow are described below.

\begin{table}[t]
    \vspace{1mm}
    \caption{
            Learning parameters of CPD in experiments
            \label{table:parameter_setting}
    }
    \vspace{-2mm}
    \begin{center}
        \begin{tabular}{@{}lp{7.5cm}llll@{}}
            \toprule
            \textbf{Para.} & \textbf{Meaning} & \textbf{Value}  \\ 
            \midrule
            $N$ & Domain division number (Pendulum) & 4 \\
             & Domain division number (the others) & 6 \\
            $m_0$ & Coefficient of MI & 1 \\
            $\gamma$ & Discount factor of RL & 0.99 \\
            $E$ & Number of episodes per domain-shift & 15 \\
            $B$ & Minibatch size (number of episodes) & 16 \\
            $T$ & Number of steps per episode & 150 \\
            \bottomrule
        \end{tabular}
    \end{center}
\end{table}

        \subsubsection{Execution Flow}
            In the first step, the previous sub-domain's local value function $Q^{(n-1)}$ is copied to current sub-domain $Q^{(n)}$. 
            At each time step $t$, local value function $Q^{(n)}$ and local policy $\pi^{(n)}$ are updated by RL through interaction with the current sub-domain. 
            To accelerate the local policy learning, we apply the monotonic policy-improvement scheme \cite{CPI,LPI,SPI} based on the similarity between the neighboring sub-domains, as assumed in the domain division step. The linear mixture composition of neighboring local policy $\pi^{(n-1)}$ and updating current local policy $\pi^{(n)}$ is addressed (details are given in the following subsection). 
            After learning in the current sub-domain for a certain number of episodes, the learning sub-domain proceeds to the next one, $n+1$. 
            The above explanation regards the forward transition starting from $n=0$ to $n=N$ and then switches to a backward transition starting from $n=N$ to $0$. These are iterated repeatedly until the learning is converged.

        \subsubsection{Policy Mixing}
            \label{MI}
            CPD approximately utilizes the MI scheme in \equref{eq:CPI-update} for accelerating the learning of local policy $\pi_{k}^{(n)}$ by appropriately exploiting neighboring sub-domain local policy $\pi^{(n')}$. This ``approximation'' will be accurate if the range of randomization parameters of the neighboring sub-domains are similar; otherwise, it will be coarse due to the domain gap. The original coefficient computation scheme in \equref{eq:CPI-update-dicide-optimal-rate} requires knowledge of the maximum reward and stationary distribution $d_{\pi,\mu}$, which is seldom possible in practice. 
            Therefore, we implement it by utilizing recent studies \cite{DCPI,CAC} that have experimentally worked well.
            As a result, we extended Eqs. (\ref{eq:CPI-update}) and (\ref{eq:CPI-update-dicide-optimal-rate}):
            \begin{align}
                \label{eq:proposed-PD-update}
                \begin{cases}
                    m_{k+1} = m_{0} \ \mathbb{E}_{s \sim \mathcal{D}^{(n)}} \big[\mathbb{E}_{a' \sim \pi^{(n')}} [Q_{k+1}^{(n)}(s,a')] \\ ~~~~~~~~~~ - \mathbb{E}_{a \sim {\pi}_{k}^{(n)}} [Q_{k+1}^{(n)}(s,{a})] \big], \\
                    \pi_{k+1}^{(n)} \gets (1-m_{k+1}) \pi_{k}^{(n)} + m_{k+1} \pi^{(n')} ,
                \end{cases}
            \end{align}
            where $m_0$ is a constant coefficient, $\mathcal{D}^{(n)}$ is the replay buffer of the current sub-domain, and $k$ is an update number. 
            Thus, the overall loss function of updating local policies $\mathcal{L}$ for learning local policy $\pi_{k}^{(n)}$ is defined:
            \begin{align}
                \label{eq:proposed-local-loss}
                \begin{cases}
                    \mathcal{L}_{\pi^{(n)}}^{\text{MI}}(\theta^{(n)}) = \text{KL}\big(\pi_{\theta^{(n)}}^{(n)}||(1-m) \pi_{\theta^{(n)}}^{(n)} + m \pi_{\theta^{(n')}}^{(n')})\big) , \\
                    \mathcal{L}_{\pi^{(n)}}(\theta^{(n)}) = \mathcal{L}_{\pi^{(n)}}^{\text{RL}}(\theta^{(n)}) + \mathcal{L}_{\pi^{(n)}}^{\text{MI}}(\theta^{(n)}) ,
                \end{cases}
            \end{align}
            where $\theta^{(n)}$ is the network parameters of the local policies, $\mathcal{L}^{\text{RL}}$ is an RL loss function, and $\mathcal{L}^{\text{MI}}$ is a loss function estimated from \equref{eq:proposed-PD-update}.

    \subsection{Global Distillation}
        At the end of the sub-domain learning, local policies $\pi^{(n)}$ are distilled into single global policy $\pi^{(g)}$, which stands in contrast to prior work that simultaneously learns both the local policies and the global one \cite{P2PDRL,DnC}. These works distill incomplete local policies in the learning process, and updating global policies repeatedly with local policy actions that offer low reward increases learning times. Thus, in CPD, learning a global policy is explicitly done only in this process after completing the learning of individual local policies. 
        
        The flow of this process is given below, where
        the aim is to learn global policy $\pi^{(g)}$ whose performance resembles those of local policies $\pi^{(n)}$ for their respective domains.
       We loop over the following three steps until reaching convergence or the maximum number of iterations:
        (1) Obtain rollout $(s_{g, t}^{(n)},a_{g, t}^{(n)}, r_{g,t}^{(n)},s_{g, t+1}^{(n)},a_{g, t+1}^{(n)} )$ in each sub-domain $n$ using $\pi^{(g)}$, since it is more sample-efficient than rollout by local policies $\pi^{(n)}$ (reason described in \cite{DistillingPD}).
        (2) Evaluate the learned local policy action probability from observation $s_{g}^{(n)}$.
        (3) Distill global policy $\pi^{(g)}$ with local policy action $a^{(n)}$.
        The above flow can be executed by optimizing the following loss function:  
        \begin{equation}
            \label{eq:proposed-Dist-marge}
            \begin{split}
            \mathcal{L}_{\pi^{(g)}}\big({\theta}\big)&=\sum_{n}\mathbb{E}_{s_{g}^{(n)},a_{g}^{(n)} \sim \mathcal{D}^{(g)}} \\
                & \ \ \bigg[
                    \text{KL}\big(\pi^{(g)}(s_{g}^{(n)},a_{g}^{(n)})||\pi^{(n)}(s_{g}^{(n)},a_{g}^{(n)})\big)
                \bigg].
            \end{split}
        \end{equation}

\begin{figure}[t]
    \centering
    \includegraphics[width=0.85\columnwidth]{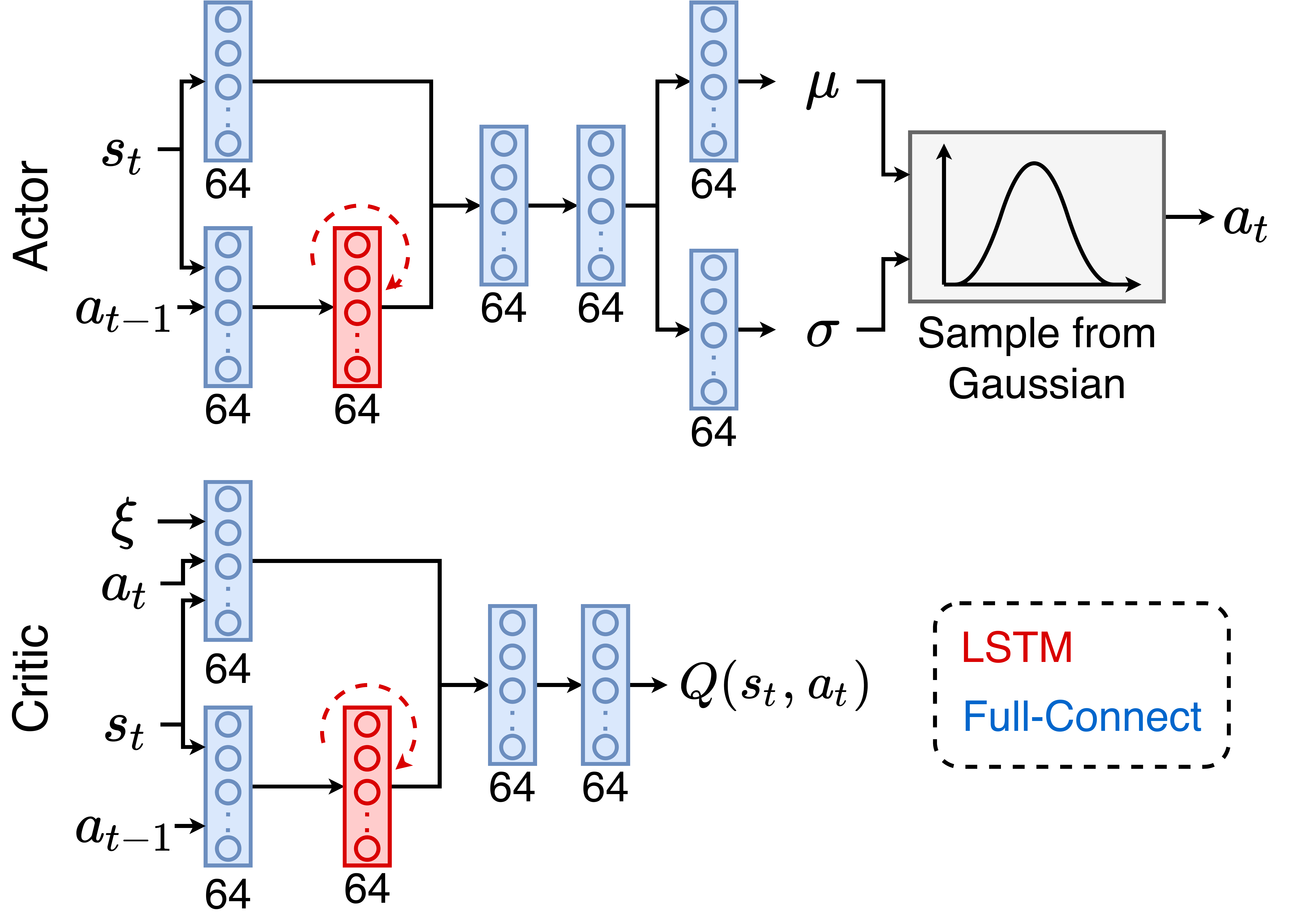}
    \caption{
    Network structure of CPD in experiment: CPD use an actor-critic structure. Actor has policy-network output that continues action $a_t$ by estimating mean $\mu$ and standard deviation $\sigma$ of Gaussian distribution used for sampling actions $a_t$.
    Critic has a value network that evaluates value function $Q$, whose inputs contain domain parameters $\xi$.
    These networks consist of fully connected layers and LSTM layers with $64$ nodes. 
    }
    \label{fig:NetworkProposed}
\end{figure}

\begin{table}
    \vspace{1.mm}
    \begin{center}
        \caption{
           Range of randomized parameters of (\textbf{Left}) Pendulum and  (\textbf{Right}) Mujoco environments
            \label{table:domain-parameter-pendulum}
            \label{table:domain-parameter-halfcheetah}
        }
        \begin{tabular}{lc}
            \toprule
                \textbf{Pendulum} & \textbf{Rate} \\
            \midrule
                \textbf{Gravity} & [0.7, 1.5] \\
                \textbf{Timestep} & [0.8, 1.2] \\
                \textbf{Bar-mass} & [0.8, 1.2] \\
                \textbf{Bar-length} & [0.8, 1.2] \\
                \textbf{Actuator-gain} & [0.7, 1.5] \\
                \textbf{Actuator-bias} & [-0.5, 0.5] \\
            \bottomrule
        \end{tabular}
        \begin{tabular}{lc}
            \toprule
                \textbf{Mujoco} & \textbf{Rate} \\
            \midrule
                \textbf{Gravity} & [0.5, 2] \\
                \textbf{Timestep} & [0.5, 2] \\
                \textbf{Friction} & [0.5, 2] \\
                \textbf{Mass} & [0.5, 2] \\
                \textbf{Actuator-gain} & [0.5, 2] \\
                \textbf{Actuator-damping} & [0.5, 2] \\
            \bottomrule
        \end{tabular}
    \end{center}
\end{table}

\section{Simulation Experiment}
\label{simulation-experimens}
    In this section, we validate the performance of the proposed CPD method in a simulation and
   investigate the following:
    \begin{enumerate}
        \item Effect of the exploitation of neighboring local policies and copying the local value functions for faster learning in CPD (Section \ref{ex-MI});
        \item Effect of the number and method of divisions of domain decomposition in CPD on learning stability and convergence (Section \ref{ex-divide});
        \item Comparison of proposed CPD method with previous works and ablation methods (Section \ref{ex-compare-sim})
    \end{enumerate}

    \subsection{Common Settings}
        \label{sim-setting}
        We leverage soft actor-critic (SAC) \cite{SAC} as the base algorithm for implementing CPD and select four tasks: OpenAI Gym Pendulum, Pusher, Swimmer, and HalfCheetah from Mujoco. The randomization settings of the domain parameters are shown in \tabref{table:domain-parameter-halfcheetah}.
        CPD has LSTM layers in both the actor and critic networks because it is virtually impossible to distinguish dynamical parameters without a recurrent network architecture or more generally a history of states and actions \cite{DR-DRL-LSTM}.
        The network architecture of CPD is shown in \figref{fig:NetworkProposed}.
        
        The actor network outputs mean $\mu$ and standard deviation $\sigma$ for a Gaussian distribution from which action $a_t$ is sampled.

\begin{figure}[t]
    \centering
    \includegraphics[width=0.99\columnwidth]{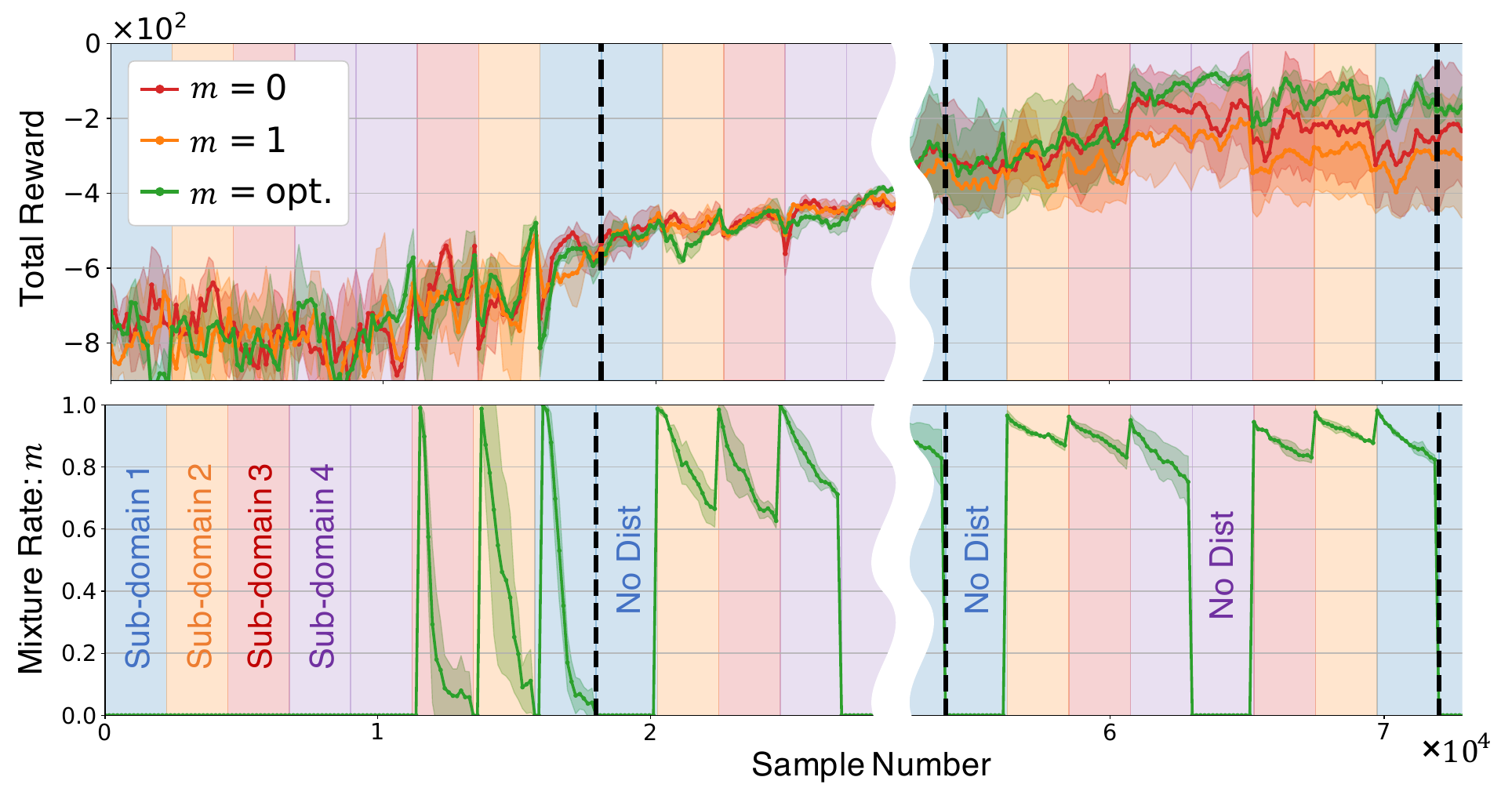}
    \caption{
        Learning curves (top) and policy mixture rate (bottom): CPD divides range of randomized parameters into four sub-domains, indicated by light-colored regions of sub-domain 1: blue; sub-domain 2: orange; sub-domain 3: red; and sub-domain 4: purple.
        CPD learns in each sub-domain and cyclically changes sub-domains by (1) forward transition and (2) backward transition.
        Dashed lines indicate one cycle of (1) and (2).
        No Dist denotes that first sub-domains of (1) and (2) are not distilled by neighboring local policies because distillations are inefficient due to neighboring policies already utilized just previously.
        $m$ denotes policy mixture rate, dynamically updated only in $m=\text{opt.}$ by \equref{eq:proposed-PD-update}.
        We started evaluating mixing rate $m$ with $120$ samples because updating of local policies starts from that number.
        Each curve is a plot of mean and variance per sample of five experiments.
    }
    \label{fig:alpha-compare}
    \label{fig:alpha-curve}
    \label{fig:evaluate-MI}
\end{figure}

    \subsection{Effect of Updating Local Policies by Mixing}
        \label{ex-MI}
        \subsubsection{Settings}
            CPD accelerates local policy learning by exploiting neighboring local policies to mix the current local policies by policy mixture rate $m$ of \equref{fig:evaluate-MI}.
            To evaluate the effectiveness of the mixing rate, we compared its performance with other baselines with constant $m=0$ or $1$ for OpenAIGym Pendulum.
            $m=0$ denotes that the neighboring local policies are not exploited for learning, and $m=1$ denotes the neighboring local policies are copied. 
            $m=\text{opt.}$ denotes optimizing $m$ by local value functions for stable acceleration of the local policy learning.

        \subsubsection{Results}
            The results are shown in \figref{fig:evaluate-MI}.
            From the total reward of \figref{fig:alpha-compare}, $m=\text{opt.}$ has the best sample efficiency and a 10 \% higher final performance than the others. The performance of $m=0$, with no acceleration by exploiting neighboring local policies, is better than $m=1$, where the neighboring local policies are exploited without optimization.
            These results show that naively exploiting neighboring local policies, i.e., $m=1$, is not always practical, perhaps due to the domain gaps. In fact, performance can be worse than without this exploitation. 

            From the transition of policy mixture rate $m$ in CPD in \figref{fig:alpha-curve}, $m$ tends to be low in the early learning stages. This indicates that exploiting neighboring local policies in the initial stage is ineffective because they are nearly random policies. As learning progresses, the local policies become more informative, and the mixing rate gains a larger quantity. Several instances where $m$ rises sharply at a certain period correspond to the sub-domain transitions' timing. As long as the learning sub-domain is fixed, $m$ decays almost monotonically, indicating that it is beneficial to smoothly shift toward using samples from the current sub-domain for better policy improvement. 

\begin{figure}[]
    \vspace{1mm}
    \centering
    \begin{tabular}{c}
        \hspace{-0.8cm}
        \begin{minipage}{0.49\linewidth}
            \centering
            \subfigure[Partitioning number]{
            \label{fig:divide}
            \includegraphics[keepaspectratio, width=0.99\linewidth, angle=0]
                        {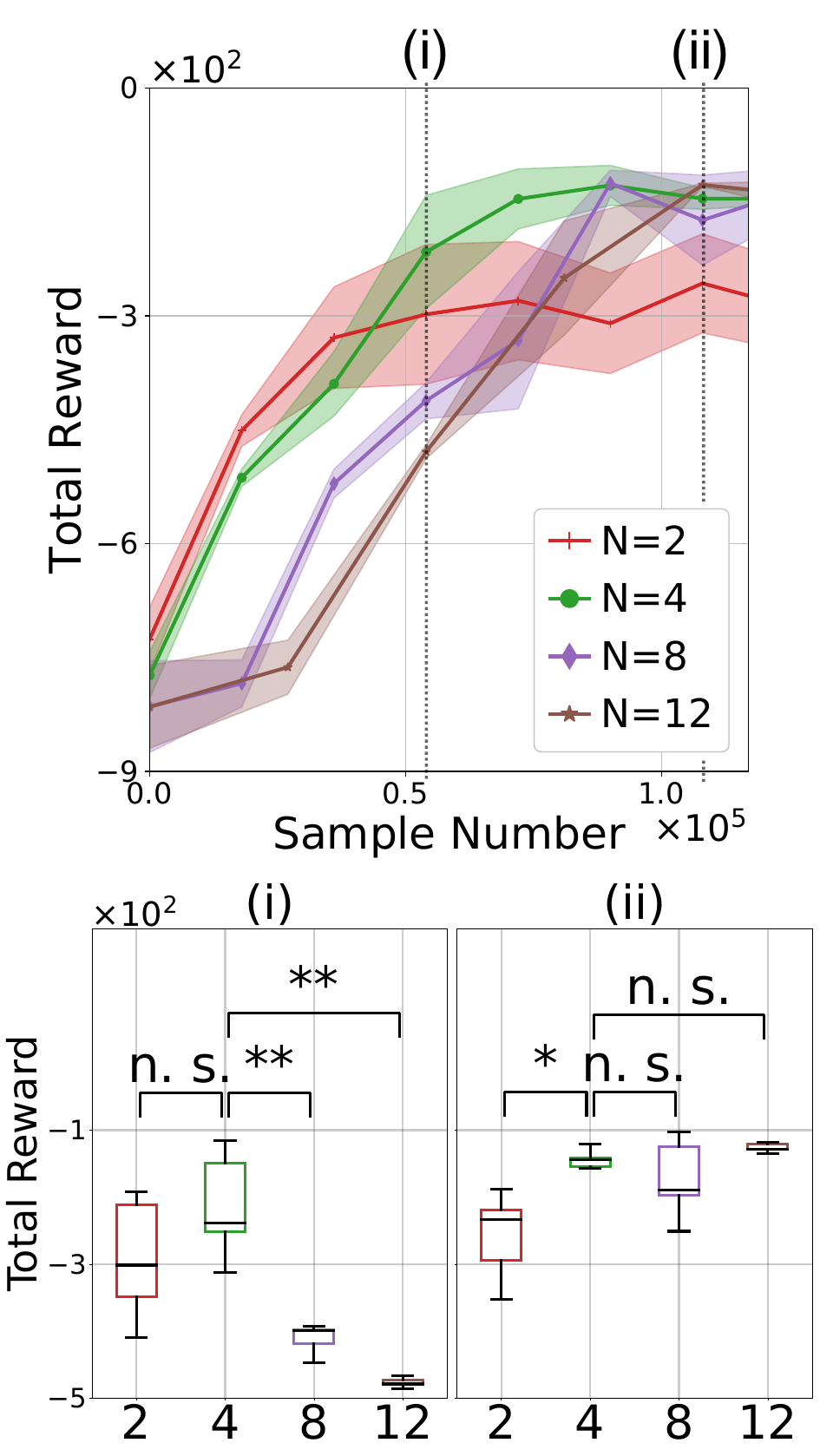}}
        \end{minipage} 
        \begin{minipage}{0.49\linewidth}
            \centering
            \subfigure[Partitioning method]{
            \label{fig:partition}
            \includegraphics[keepaspectratio, width=0.99\linewidth, angle=0]
                        {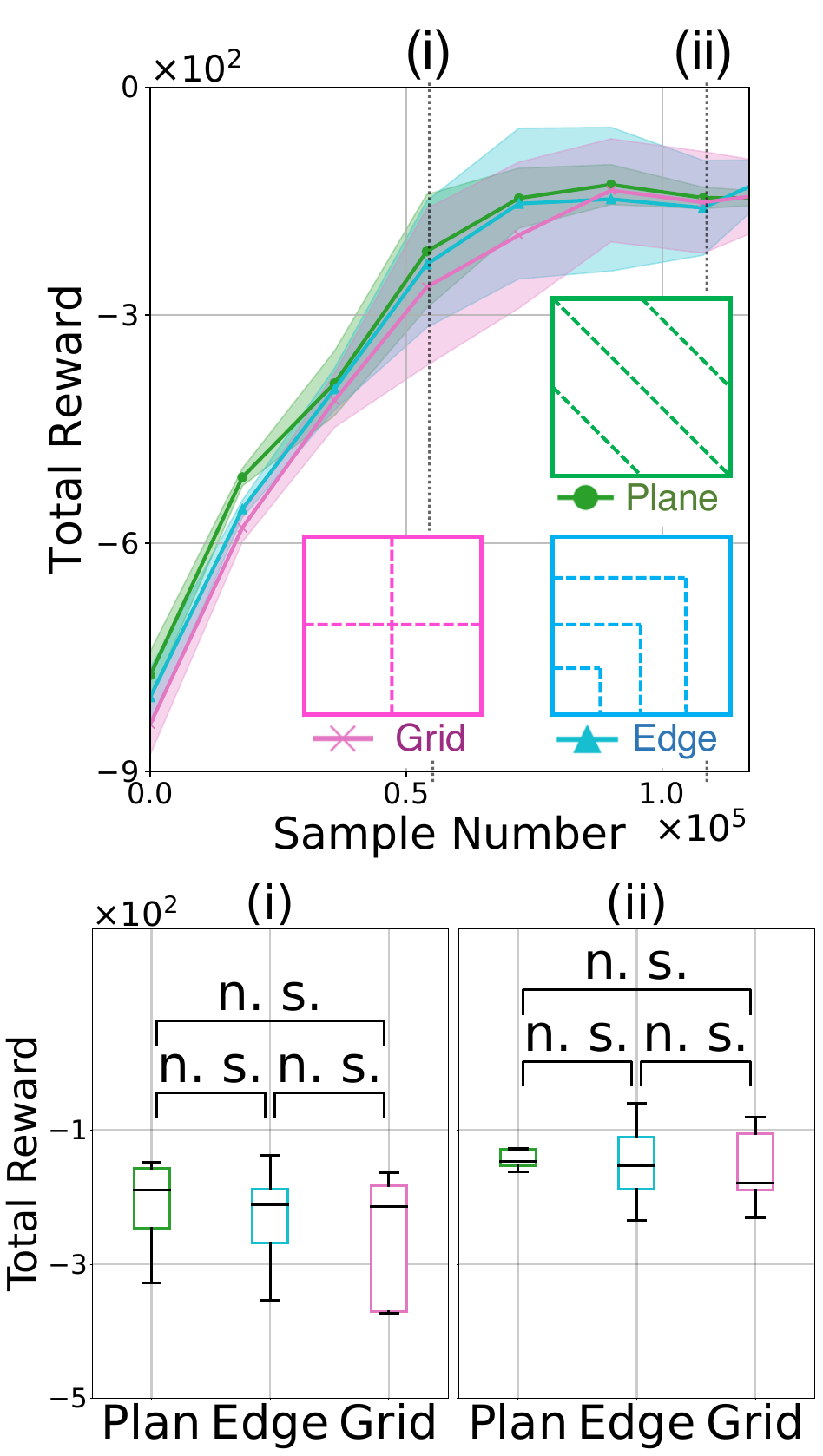}}
        \end{minipage} 
    \end{tabular}
    \caption{
        Performance comparison: (a) different numbers of partitions of randomized parameter range, (b) different patterns of partitioning methods.
        Partitioning methods are evaluated in three types, plane, edge, and grid, as shown in (b). 
        These outer lines indicate the entire range of parameters. Each vertical and horizontal range indicates the value of one parameter. Dashed lines indicate how to divide parameters. All results of the partitioning method are evaluated after splitting four.
        Each curve of (a) and (b) plots the mean and variance per sample over five experiments.
        The bottom boxplots are evaluated in (i) $54,000$ samples and (ii) $108,000$ samples of each curves (\textasteriskcentered \ and \textasteriskcentered\textasteriskcentered \ mean $p < 0.05$ and $p < 0.005$).
    }
    \label{fig:divide-partition}
\end{figure}

\begin{figure}
    \vspace{1mm}
    \centering
    \begin{tabular}{c}
        \hspace{-0.6cm}
        \begin{minipage}{0.47\linewidth}
            \centering
            \subfigure[Pendulum]{
            \label{fig:compare_others_pendulum}
            \includegraphics[keepaspectratio, width=0.99\linewidth, angle=0]
                        {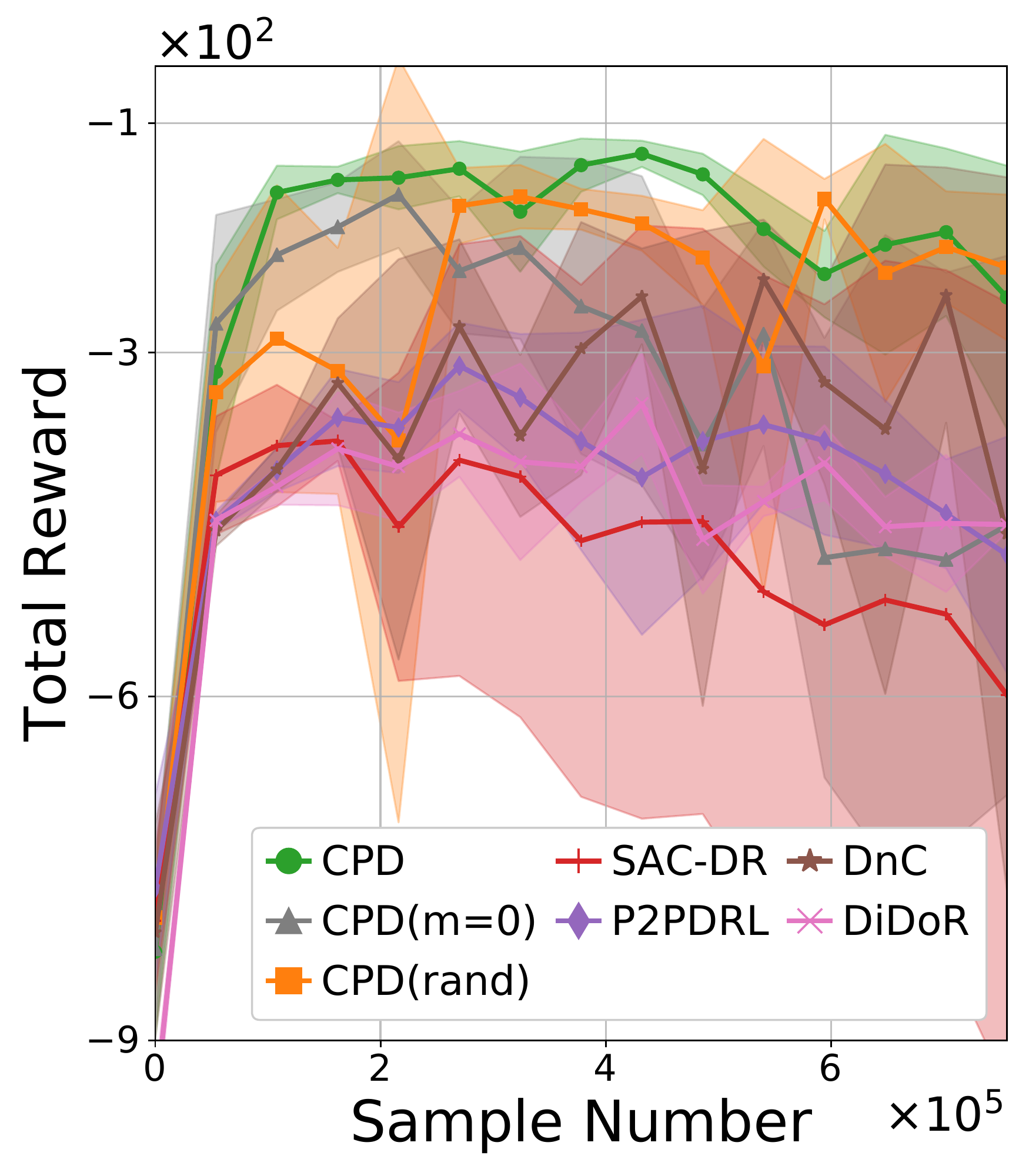}}
        \end{minipage} 
        \begin{minipage}{0.47\linewidth}
            \centering
            \subfigure[Pusher]{
            \label{fig:compare_others_pusher}
            \includegraphics[keepaspectratio, width=0.99\linewidth, angle=0]
                        {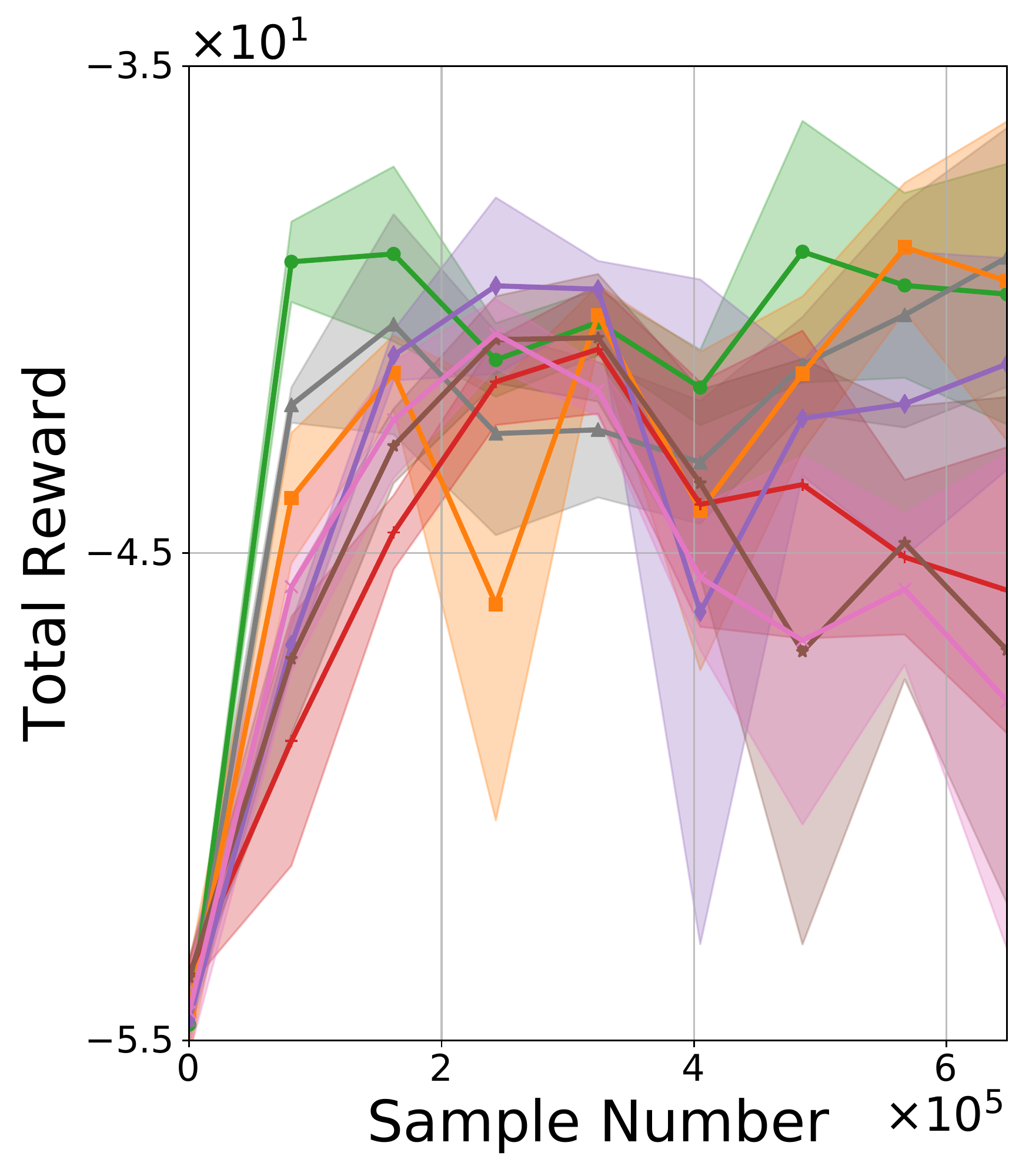}}
        \end{minipage} 
        \\
        \hspace{-0.6cm}
        \begin{minipage}{0.49\linewidth}
            \centering
            \subfigure[Swimmer]{
            \label{fig:compare_others_swimmer}
            \includegraphics[keepaspectratio, width=0.99\linewidth, angle=0]
                        {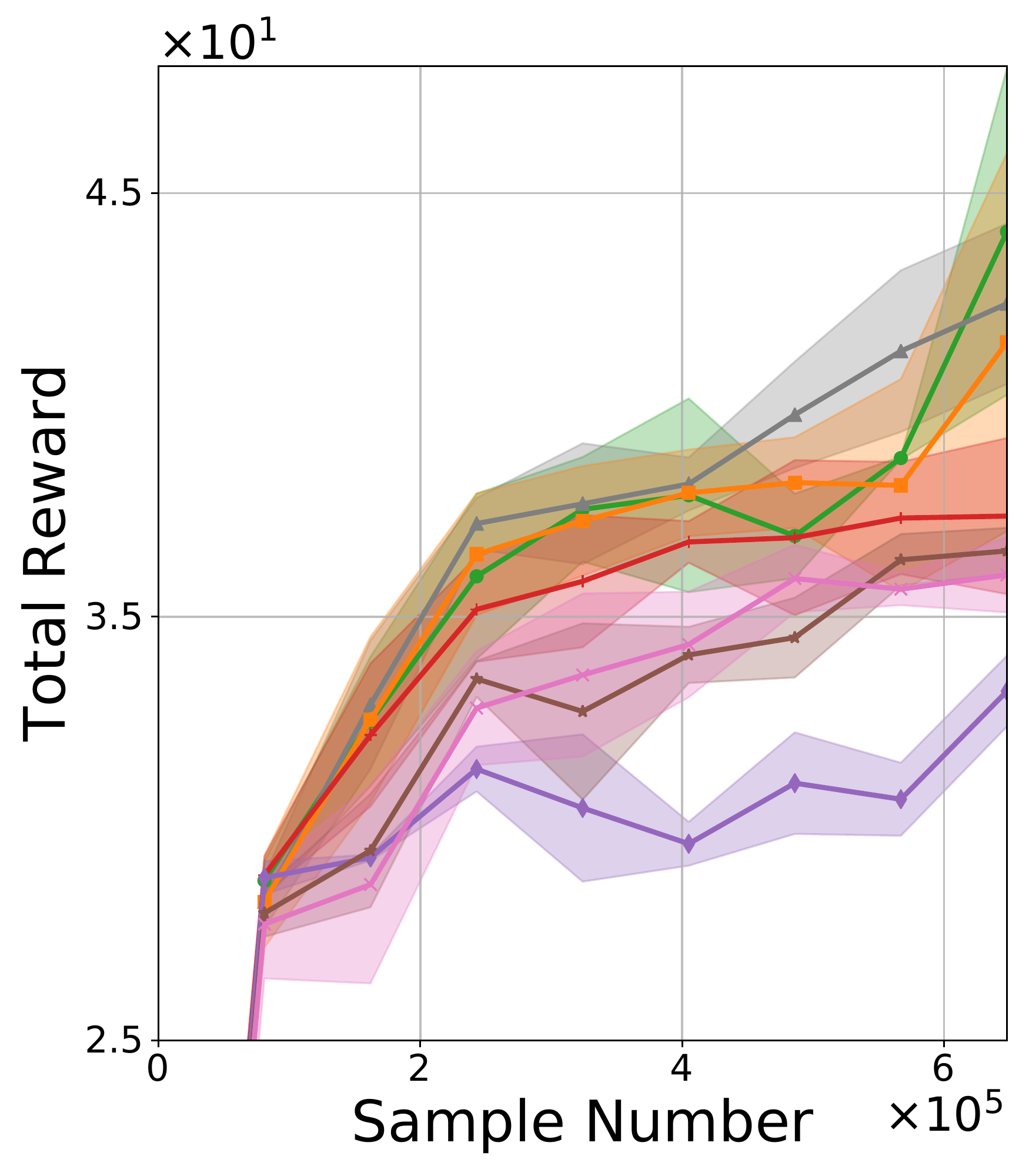}}
        \end{minipage} 
        \begin{minipage}{0.49\linewidth}
            \centering
            \subfigure[HalfCheetah]{
            \label{fig:compare_others_halfcheetah}
            \includegraphics[keepaspectratio, width=0.99\linewidth, angle=0]
                        {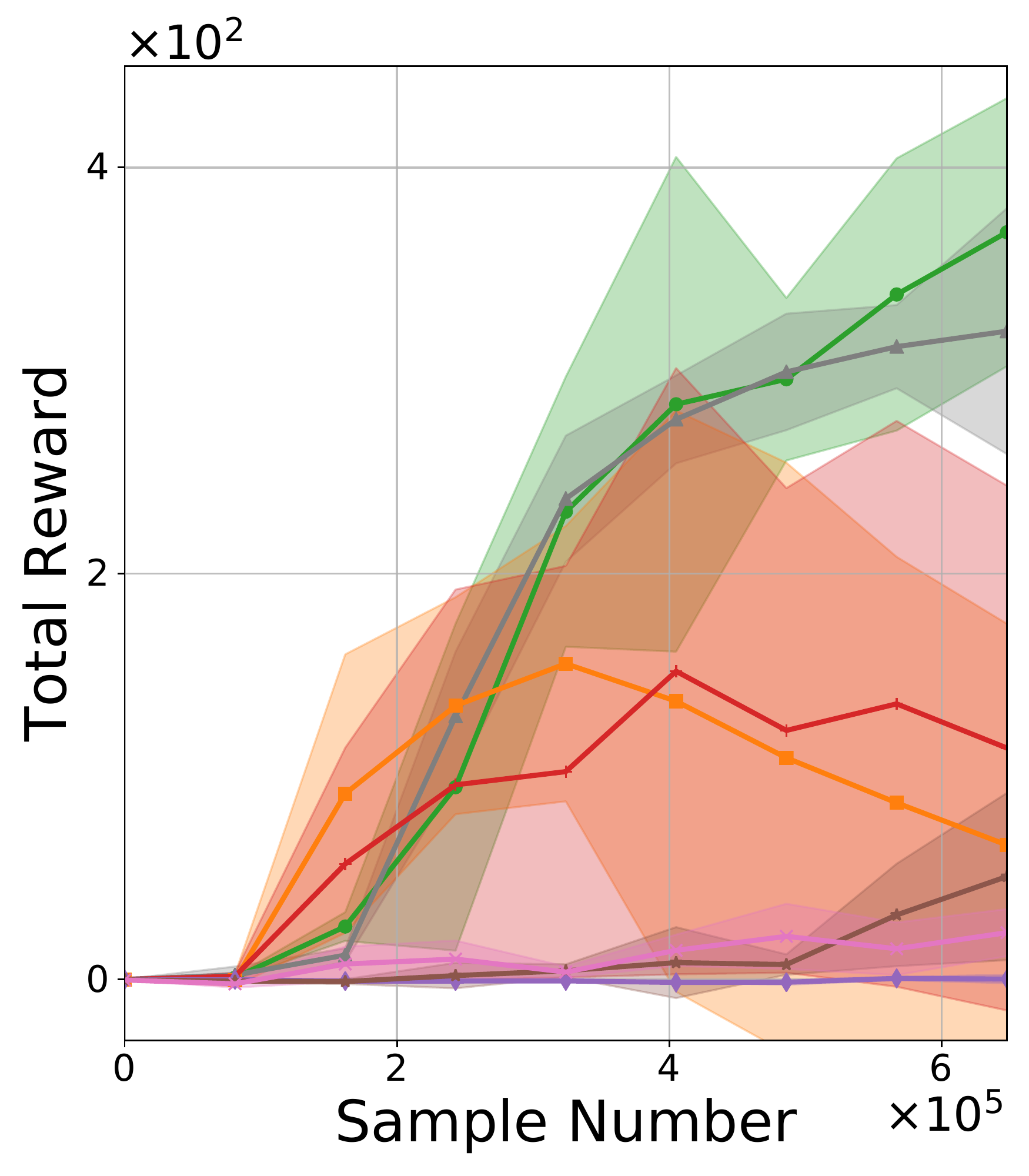}}
        \end{minipage} 
    \end{tabular}
    \caption{
        Learning curves of (a) Pendulum, (b) Pusher, (c) Swimmer, and (d) HalfCheetah for different methods:
        SAC-DR is SAC using single-policy learning with DR, and P2PDRL, DnC, and DiDoR are multiple-policy learning methods with range of randomized parameters divided, as described in Section \ref{stable-DR}.
        CPD, CPD ($m=0$), and CPD (rand) respectively indicate CPD with full components, CPD without exploiting neighboring local policies, and CPD with random transitioning of sub-domains.
        Each curve plots mean and variance over five experiments.
    }
    \label{fig:sim_result-all}
\end{figure}
    
    \subsection{Evaluation of Number and Method of Divided Ranges}
        \label{ex-divide}
        \subsubsection{Settings}
            Since CPD divides the range of the randomized parameters into $N$ sub-domains for learning local policies, $N$ and the partitioning method may influence the final performance and sample efficiency. 
            Thus, we investigated the CPD performance with various $N$ and partitioning methods on Pendulum.

        \subsubsection{Results}
            As shown in \figref{fig:divide} shows that CPD partitioning the parameters in $N \geq 4$ reached comparable performance without a significant difference at (ii) 108,000 samples. On the other hand, at (i) 54,000 samples, meaning half samples of (ii), the performances of $N \leq 4$ are higher than $N > 4$ with a significant difference. The results indicate a trade-off in $N$ between final performance and sample efficiency. The optimal balance is found at $N=4$, and increasing $N$ above that value gradually decreases sample efficiency. Based on this result, all subsequent experiments were evaluated by $N=4$ due to the best sample efficiency and performance.
            
            Although a partitioning method of CPD can be freely designed, performances might differ. In \figref{fig:partition}, we evaluated the performance differences in three types of partitioning methods: plane, splitting so that there is one adjacent region for each axis; grid, splitting into blocks for each axis; and edge, splitting at right angles to each axis. As shown in \figref{fig:partition}, the proposed method is robust to partitioning patterns in terms of total reward and sample efficiency with no significant difference at (i) and (ii). Based on this result, all subsequent experiments were evaluated by the plane method.

    \subsection{Comparison with other Methods}
        \label{ex-compare-sim}
        \subsubsection{Settings}
            As a common setting of CPD and other baselines, SAC was used as a base-RL method. The difference between CPD and the baselines is in the distillation process of the local and global policies, details of which are shown as follows and summarized in \tabref{table:calculation-performance}: 
            \begin{quote}
            \begin{itemize}
                \item CPD updates the current local policies by the distillation of one neighboring policy per step. It finally mixes local policies into a global policy.
                \item CPD, CPD ($m=0$), and CPD (rand) respectively indicate CPD with full components, CPD without exploiting neighboring local policies, and CPD with random transitioning of sub-domains. CPD ($m=0$) and CPD (rand) are the same update count as the original CPD.
                \item SAC-DR updates the current policy only by RL per step. SAC-DR never mixes a policy and finally uses the current policy as a global policy.
                \item P2PDRL updates the current local policies by mutual distillation between all the local policies in total ${}_N {\rm P}_2$ update per step. P2PDRL never mixes a policy and finally uses one of the local policies as a global policy. 
                \item DnC updates the current local policies only by RL per step. At each end of the iterations, DnC mixes the local policies into a global policy.
                \item DiDoR updates current local policies only by RL per step. DiDoR finally mixes local policies into a global policy.
            \end{itemize}
            \end{quote}
            We conducted experiments with four environments: Pendulum, Pusher, Swimmer, and HalfCheetah. 
            In this section, we evaluate the learning performance by the rewards and the cost by the learning schedule and learning time.

        \subsubsection{Results of Learning Performance}
            As shown in \figref{fig:compare_others_pendulum}, CPD shows the highest performance by achieving six- and three-times better sample efficiencies than the methods that outperformed all the others. SAC, P2PDRL, and DiDoR only achieved a low total reward that was under $70 \%$ of CPD's. 
            As shown in \figref{fig:compare_others_halfcheetah}, none of the methods (except CPD) learned effective policies with a practical number of samples. SAC-DR, which does not partition the range of randomization parameters, learns quickly; however, its highest total reward is only about $40 \%$ of CPD. 
            For the other environments shown in \figref{fig:compare_others_pusher} and \figref{fig:compare_others_swimmer}, CPD achieved the highest total reward and sample efficiency.
            
            We also applied CPD ($m=0$) and CPD (rand) as an ablation study. 
            For Pendulum, CPD (rand) achieved about a three times better sample efficiency than CPD ($m=0$), reaching a maximum total reward of CPD ($m=0$). However, CPD ($m=0$) suffered a significant performance degradation after learning convergence.
            Furthermore, CPD (rand) showed less sample efficiency than CPD ($m=0$).
            
            In summary, our comparisons to previous works show the higher sample efficiency of the proposed method, and ablation studies show the effectiveness of the proposed architecture with a monotonic policy-improvement scheme and neighboring transitions of the sub-domain. 

\begin{table}[t]
    \caption{
            Learning cost for each algorithm:
            Dist./Step denotes policy updates by distillation per step.
            Mix/Iteration and Mix/End denote policy mixing from local policies to a global policy per iteration and end of learning.
            Time denotes learning time of 1000 episodes in Pendulum task, evaluated in mean per $5$ experiments.
            Time is measured by Intel Core i9-9900X and Nvidia RTX-3080Ti.
            $N$ denotes number of local policies.
            \label{table:calculation-performance}
    }
    \vspace{-2mm}
    \begin{center}
        \begin{tabular}{lccccc}
            \toprule
            \textbf{Method} & \textbf{Dist./Step} & \textbf{Mix/Iteration} & \textbf{Mix/End} & \textbf{Time [min.]}\\ 
            \midrule
            \textbf{CPD} & $1$ & - & \checkmark & $107$ \\
            \textbf{SAC-DR} & $0$ & - & - & $86$ \\
            \textbf{P2PDRL} & ${}_{N} {\rm{P}}_2$ & - & - & $139$ \\
            \textbf{DnC} & $0$ & \checkmark & \checkmark & $120$ \\
            \textbf{DiDoR} & $0$ & - & \checkmark & $92$ \\
            \bottomrule
        \end{tabular}
    \end{center}
\end{table}

        \subsubsection{Results of Learning Cost}
            As shown in \tabref{table:calculation-performance}, the fastest algorithm is SAC-DR because it has no distillation. CPD has a moderate learning time due to some policy mixing steps: $24 \%$ lower than SAC-DR and $30 \%$ faster than P2PDRL.

\begin{table}
    \vspace{1.mm}
    \begin{center}
    \caption{
        Range of randomized parameters in ball-dispersal task: 
        Ball-place denotes center position of piled balls.
        When true ball-position is $(x,y,z)$,
        randomized positions are $(x\!+\!{\rm X_\text{bias}}, y\!+\!{\rm Y_\text{bias}}, z\!\times\!{\rm Z_\text{weight}}\!+\!{\rm Z_\text{bias}})$.
        \label{table:domain-parameter-real}
    }
        \begin{tabular}{lcclcc}
            \toprule
                \textbf{Physical} & \textbf{min} & \textbf{max} & \textbf{Observation} & \textbf{min} & \textbf{max} \\
            \midrule
                \textbf{Gravity} $[\rm{m/s^2}]$ & 9 & 11 & $\bf{X_\text{bias}}$ $[\rm{mm}]$ & -10 & 10 \\
                \textbf{Friction-coeff.} $[\cdot]$ & 0.6 & 1 & $\bf{Y_\text{bias}}$ $[\rm{mm}]$ & -10 & 10 \\
                \textbf{Ball-mass} $[\rm{g}]$ & 5 & 20 & $\bf{Z_\text{bias}}$ $[\rm{mm}]$ & 0 & 10 \\
                \textbf{Ball-place} $[\rm{mm}]$ & -125 & 125 & $\bf{Z_\text{weight}}$ $[\cdot]$ & 0.7 & 1.3 \\
                \textbf{Ball-radius} $[\rm{mm}]$ & 25 & 30 &  & & \\
            \bottomrule
        \end{tabular}
    \end{center}
\end{table}

\section{Real-Robot Experiment}
    \label{real-experiments}
    In this section, we verify the effectiveness of the proposed method (CPD) on a ball-dispersal task as a real-world task (\figref{fig:senzai}).
    As shown in \figref{fig:senzai}, the upper three rows show the randomized simulation environment, and the bottom row shows the real-world environment.
    This task is characterized by the countless ball-placement patterns that are possible in the process of knocking down a pile of many balls.
    Thus, the agent must efficiently learn an effective policy that can manage such innumerable ball patterns. 
    This task is appropriate for a sim-to-real approach because it requires that the balls are restacked each time and that many initializations are made in the real-robot environment.

\begin{figure}[t]
    \centering
    \includegraphics[width=0.8\columnwidth]{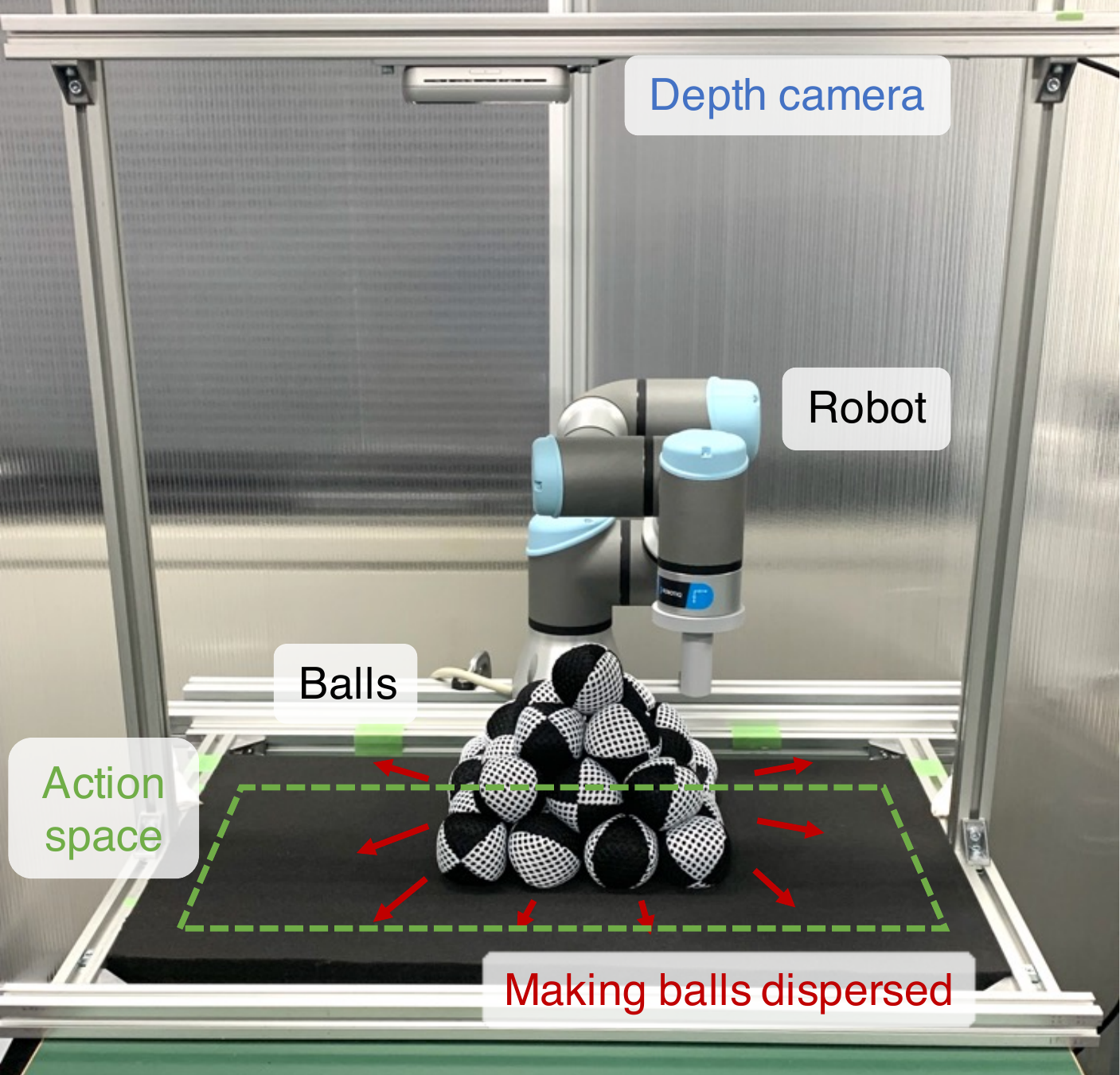}
    \caption{
    Experiment settings: 
    It uses UR3, and task goal is for robot hand's to knockdown the balls.
    Robot's hand can move in action space shown as a green dashed line.
    Depth camera captures action space for getting observations.
    }
    \label{fig:experimet_settings_real}
\end{figure}

    \subsection{Settings}
        The settings of the real-robot experiment are shown in \figref{fig:experimet_settings_real}. 
        Given depth image $s^{\text{depth}}$ obtained from the depth camera mounted above the robot and 3D coordinates $s^{\text{position}}$ of the robot's hand, the observation is defined as $s=[\text{Conv}(\text{Conv}(s^{\text{depth}})), s^{\text{position}}]$.
        The robot initializes its position for each action $a$ to obtain the balls' depth image without occlusion. 
        The robot hand's rotation is fixed so that it always faces the floor.
        $\text{Conv}$ is a convolution operation with a kernel size of $3$. $s^{\text{depth}}$ compresses the number of pixels from $[480,640]$ to $[5,9]$.
        $s^{\text{depth}}$ is visualized in the second and fourth lines of \figref{fig:ex-time-series}.
        Action $a$ specifies the robot's fingertip in 3D relative coordinates and moves it from the coordinates of the previous action.
        The distance that can be moved in one action is limited to 30 mm for each coordinate axis.
        Reward $r$ is determined from the balls' height to evaluate their collapse.
        The number of balls is $B=30$; the reward is $r=\sum_{b=1}^{B} -h_i$, 
        where $h_i$ is the height of the $i$-th ball. 
        The goal of this task is to make all the balls lay flat on the table. In other words, create a state in which no balls are on top of any other balls.
        
        Only the robot's fingertip is simulated in the simulation environment because, in a real-world environment, the robot's kinematics do not affect the task achievement.
        Observations $s$ are formatted identically as in the real-robot because depth information can be accurately obtained in the simulation, and $s^{\text{depth}}$ is directly assigned as the simulation ball-height to each pixel without initializing the robot position.
        The definitions of actions $a$ and rewards $r$ are identical as in the real-world environment. 
        The domain parameters used for DR and their ranges are shown in \tabref{table:domain-parameter-real}.
        This task randomizes the physical and sensor model's parameters.
        Since balls in the real world are deformed by the robot's physical interaction, making them move in strange directions, parameters including ball-radius and friction-coefficient are randomized to express this phenomenon. Furthermore, the depth sensor is so noisy that observations of depth images between simulations and the real world have huge gaps, and thus observations of the ball positions and sizes are randomized by injecting noise.
        SAC updates the local policies and local value functions as a base-RL method with the same networks and setting as in Section \ref{sim-setting}.
        Due to the greater task difficulty compared to the simulation experiments, all the node numbers of networks were increased to $128$.

\begin{figure}[t]
    \centering
    \includegraphics[width=0.9\columnwidth]{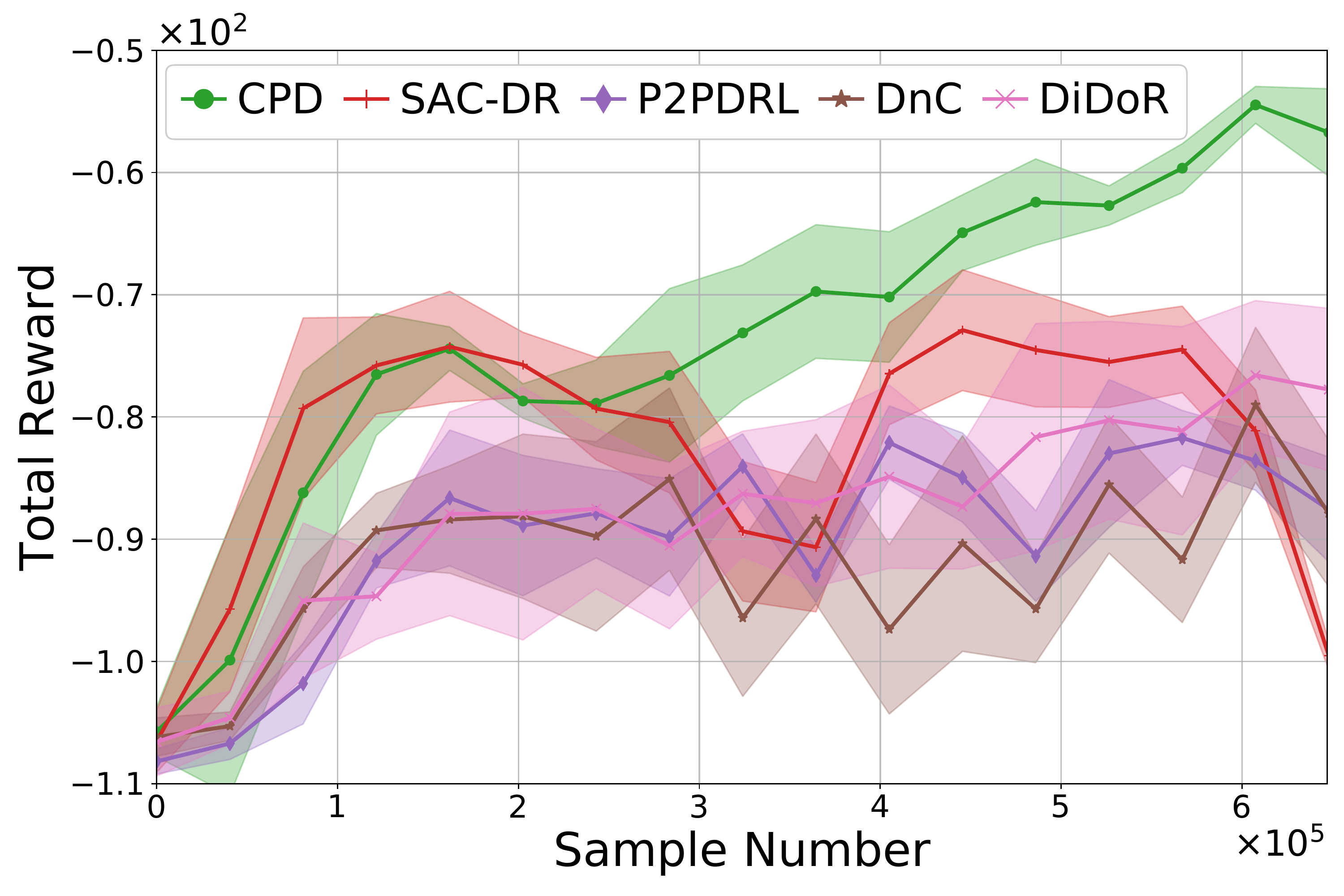}
    \caption{
    Learning curves of ball-dispersal task:
    Each curve plots mean and variance of total reward over five experiments.
    }
    \label{fig:compare_others_ball}
\end{figure}

    \subsection{Results}
        CPD and other baselines are compared in \figref{fig:compare_others_ball}.
        SAC-DR converges faster than CPD in the initial stage of learning, although its performance soon becomes unstable.
        The top total reward of SAC-DR can be achieved by one touch of the balls, which then transitions to an enormous number of patterns, and so SAC-DR cannot identify good actions under such conditions.   
        On the other hand, CPD shows stable performance improvement and achieves around $40 \%$ better total reward than the other baselines.
        
        To evaluate the effectiveness of the learned policies for a sim-to-real transfer, we prepared two test environments with different types of balls: juggling and bead. 
        \tabref{table:achievement} shows the results of evaluating how well the balls are knocked down using the learned policies from the experiments shown in \figref{fig:compare_others_ball}.
        We also evaluated SAC learned without DR as a comparison. 
        Just as SAC-DR failed to achieve good performance in simulations, the method only knocked down about half of the balls in the real-world environment of juggling balls. Furthermore, SAC-DR failed to knock the bead balls down, an action that is more difficult than for the juggling balls. In fact, SAC could not even reach the balls.
        In contrast, CPD knocked down almost all of both types of balls. 

\begin{figure}
    \hspace{-3.6mm}
    \centering
    \includegraphics[width=0.99\columnwidth]{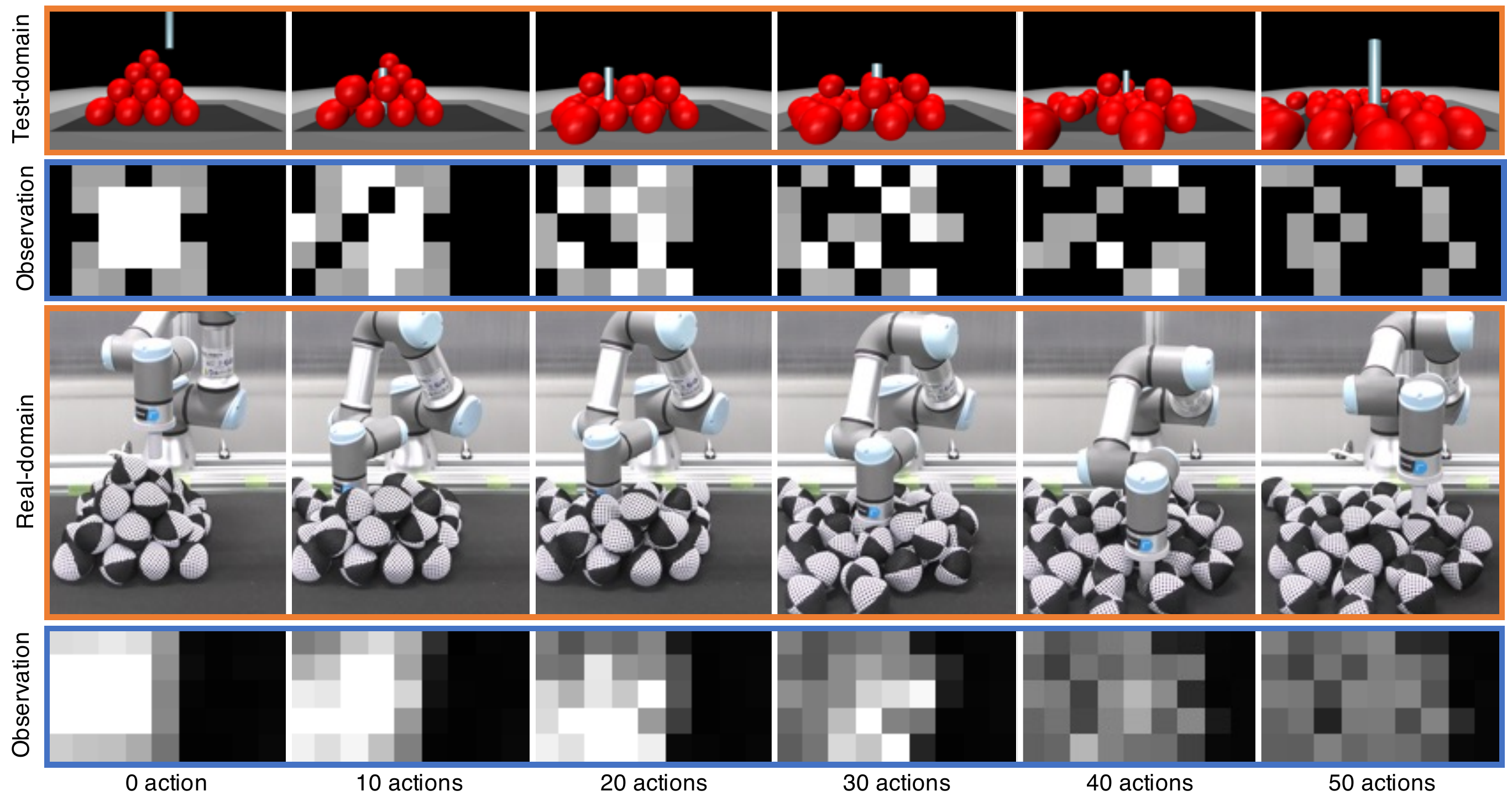}
    \caption{
        Snapshots of robot's state and observation in ball-dispersal task executed by CPD's learned policy:
        Upper two rows are environments and observations of simulation. Lower two rows are environments and observations of real world.
        These observations are depth images compressed to $(5,9)$.
        Simulation observations are randomized by DR as given in \tabref{table:domain-parameter-real}. 
    }
    \label{fig:ex-time-series}
\end{figure}

        \figref{fig:ex-time-series} shows snapshots of the simulation and real-world environments with the learned CPD policy experimentally obtained shown in \figref{fig:compare_others_ball}.
        From the observation, the scale and noise of the balls' position and depth information are different between the simulation and the real world.
        However, since the scale and noise of the balls' position and depth information are widely randomized, the task is achieved by transferring the simulation policies to the real world in a zero-shot setting.

\begin{table}[t]
    \begin{center}
        \caption{
            Task achievement of learned policies of ball-dispersal task: Final-view denotes final result of evaluating policy actions in one episode. Achieve represents task achievement, assessed by number of balls knocked to floor.
            This experiment uses two types of balls: (1) juggling balls that are hard and easy to move and (2) bead balls that are easily deformed and difficult to move.
            This experiment used 30 balls: 16 first placed on the floor and other 14 balls on top of the first 16 balls. Second set of 14 balls is used to evaluate task achievement.
            Each value of task achievement is a mean of over five experiments per learned policy.
        }
        \label{table:achievement}
        \begin{tabular}{cccc}
            \toprule
                \textbf{Method} & \textbf{SAC} & \textbf{SAC--DR} &  \textbf{CPD} \\
            \midrule
                \parbox{10mm}{\textbf{Final view (juggling)}} &
                \begin{minipage}{0.22\columnwidth}
                  \scalebox{0.39}{\includegraphics{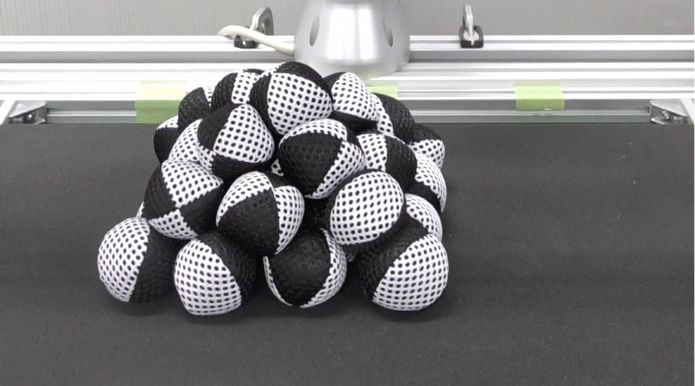}}
                \end{minipage} &
                \begin{minipage}{0.22\columnwidth}
                  \scalebox{0.39}{\includegraphics{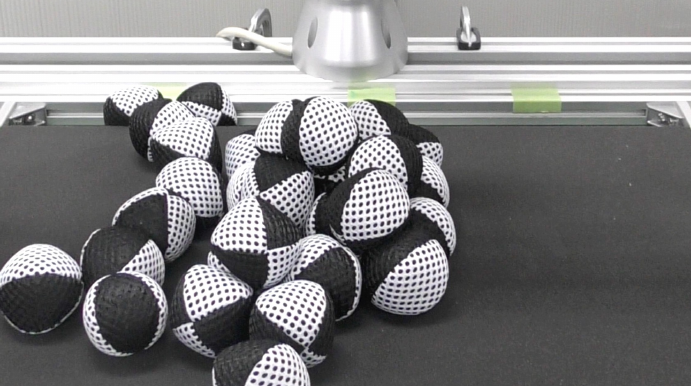}}
                \end{minipage} & 
                \begin{minipage}{0.22\columnwidth}
                  \scalebox{0.39}{\includegraphics{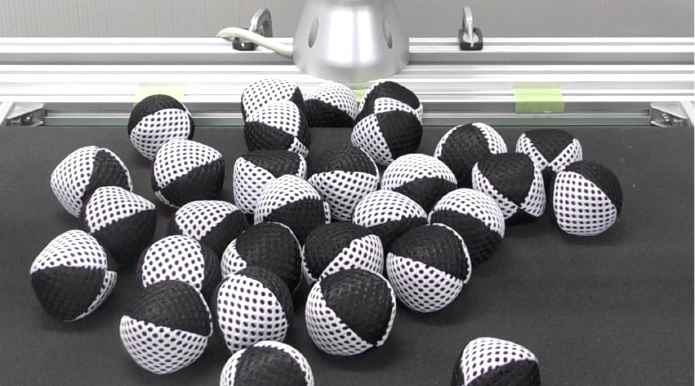}}
                \end{minipage} \\
            \midrule
                \parbox{10mm}{\textbf{Achievement}} & 0 / 14  & 6 / 14 &  13 / 14 \\
            \midrule
                \parbox{10mm}{\textbf{Final view (beads)}} &
                \begin{minipage}{0.22\columnwidth}
                  \scalebox{0.33}{\includegraphics{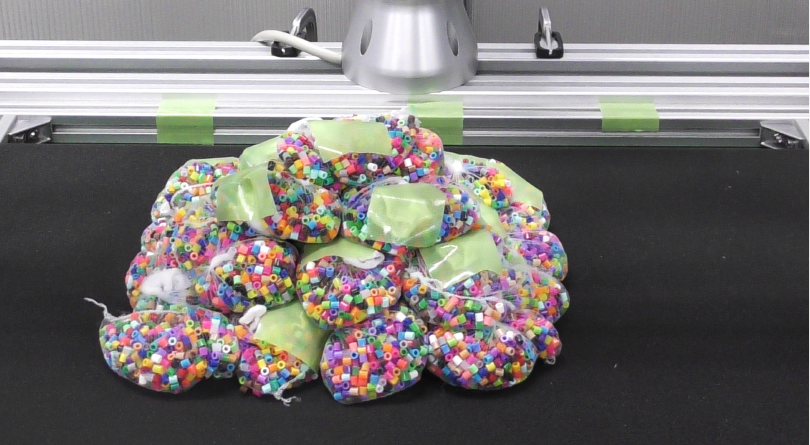}}
                \end{minipage} &
                \begin{minipage}{0.22\columnwidth}
                  \scalebox{0.33}{\includegraphics{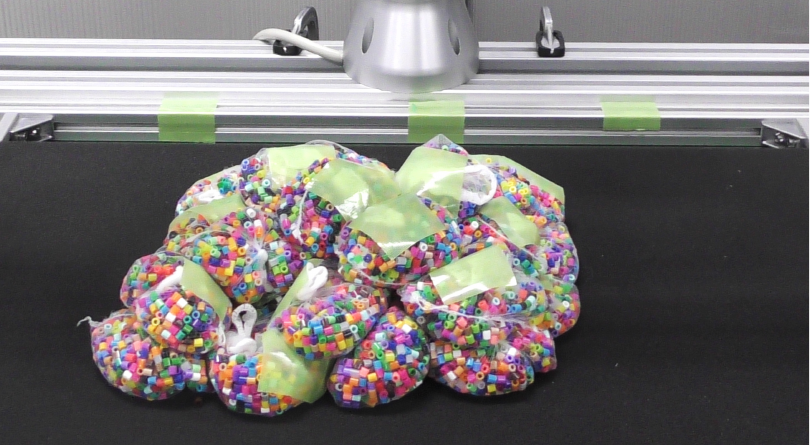}}
                \end{minipage} & 
                \begin{minipage}{0.22\columnwidth}
                  \scalebox{0.33}{\includegraphics{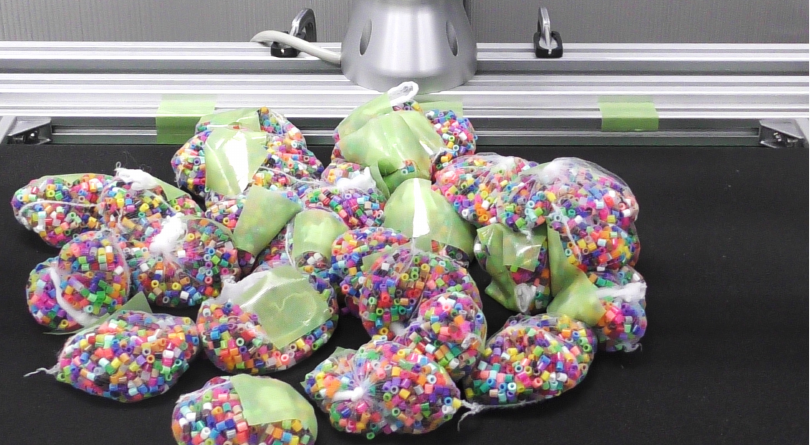}}
                \end{minipage} \\
            \midrule
                \parbox{10mm}{\textbf{Achievement}} & 0 / 14  & 1 / 14 &  11 / 14 \\
            \bottomrule
        \end{tabular}
    \end{center}
\end{table}

\section{Discussions}
    \label{discussion}
    In the current CPD framework, dividing number $N$ of the range of randomized parameters needs to be set manually. As described in Section \ref{ex-divide}, dividing number $N$ controls the tradeoff between the learning performance and the sample efficiency. 
    We addressed three different partitioning patterns for the proposed method and confirmed their robustness to these patterns. In common, they gave equal importance to all the sub-domains, although they may not all contribute equally to policy distillation. Future work will develop a mechanism to identify sub-domains that do not contribute to learning and exclude them from periodic transitions.

    In the current implementation, the decision regarding which parameters to split is determined by the designer's knowledge. Parameters that significantly change the reward scale or the behavior of the actor depending on the parameter should be selected and split. Exploring its automatic approach is also future work. 
    
    Since CPD learns by cyclically transitioning through each sub-domain, it does not seem straightforward to extend it to a parallelized RL \cite{Gorila,A3C,Ape-X}.  However, CPD can indeed be extended to a parallelized RL by grouping a certain number of sub-domains into one cluster. Such a method could be applied to larger domains.
    
    Since our proposed method has an actor-critic structure for RL, it requires policies and action value functions. Applicable learning algorithms include the latest DRL algorithms, such as SAC, TD3 \cite{TD3}, and DDPG \cite{DDPG}. This paper utilizes SAC, which shows the best performance of the actor-critic method \cite{SAC}.
    
    The experimental results indicated that our CPD empirically achieved better sample efficiency than the previous methods for multiple tasks. However, its theoretical justification remains open since the theoretical guarantees of the original MI method were obtained assuming a single domain \cite{CPI} and might not hold in a randomized environment.

\section{Conclusion}
    \label{conclusion}
    The following are this paper's main contributions: 1) it proposed CPD, a new DRL method that is sample-efficient with DR in a zero-shot setting; 2) it utilized a monotonic policy-improvement scheme proposed in single-domain RL setting to a multi-domain RL setting and successfully sped-up the policy performance increase; 3) it made a numerous ball environment in both simulations and the real world as preliminary tasks of sand and powder operations and achieved its task in the zero-shot setting.

\section*{Acknowledgments}
    This work was supported by JST Moonshot Research and Development, Grant Number JPMJMS2032. This work was partially supported by the JSPS KAKENHI, Grant Number JP21H03522.

\bibliography{paper}

\begin{thebibliography}{10}
\expandafter\ifx\csname url\endcsname\relax
  \def\url#1{\texttt{#1}}\fi
\expandafter\ifx\csname urlprefix\endcsname\relax\def\urlprefix{URL }\fi
\expandafter\ifx\csname href\endcsname\relax
  \def\href#1#2{#2} \def\path#1{#1}\fi

\bibitem{dqn}
V.~Mnih, K.~Kavukcuoglu, D.~Silver, A.~A. Rusu, J.~Veness, M.~G. Bellemare,
  A.~Graves, M.~Riedmiller, A.~K. Fidjeland, G.~Ostrovski, et~al., Human-level
  control through deep reinforcement learning, Nature 518~(7540) (2015)
  529--533.

\bibitem{google-picking}
S.~Levine, P.~Pastor, A.~Krizhevsky, J.~Ibarz, D.~Quillen, Learning hand-eye
  coordination for robotic grasping with deep learning and large-scale data
  collection, The International Journal of Robotics Research (IJRR) 37~(4-5)
  (2018) 421--436.

\bibitem{drl-door}
S.~Gu, E.~Holly, T.~Lillicrap, S.~Levine, Deep reinforcement learning for
  robotic manipulation with asynchronous off-policy updates, in: IEEE
  International Conference on Robotics and Automation (ICRA), 2017, pp.
  3389--3396.

\bibitem{DR-origin}
J.~Tobin, R.~Fong, A.~Ray, J.~Schneider, W.~Zaremba, P.~Abbeel, Domain
  randomization for transferring deep neural networks from simulation to the
  real world, in: IEEE/RSJ International Conference on Intelligent Robots and
  Systems (IROS), 2017, pp. 23--30.

\bibitem{DR-deformable}
J.~Matas, S.~James, A.~J. Davison, Sim-to-real reinforcement learning for
  deformable object manipulation, in: Conference on Robot Learning (CoRL),
  2018, pp. 734--743.

\bibitem{ActiveDR}
B.~Mehta, M.~Diaz, F.~Golemo, C.~J. Pal, L.~Paull, Active domain randomization,
  in: Conference on Robot Learning (CoRL), 2020, pp. 1162--1176.

\bibitem{P2PDRL}
C.~Zhao, T.~Hospedales, Robust domain randomised reinforcement learning through
  peer-to-peer distillation, in: Asian Conference on Machine Learning (ACML),
  2021, pp. 1237--1252.

\bibitem{NeuralSim}
E.~Heiden, D.~Millard, E.~Coumans, Y.~Sheng, G.~S. Sukhatme, Neuralsim:
  Augmenting differentiable simulators with neural networks, in: IEEE
  International Conference on Robotics and Automation (ICRA), 2021, pp.
  9474--9481.

\bibitem{sim2real-si}
W.~Yu, V.~C. Kumar, G.~Turk, C.~K. Liu, Sim-to-real transfer for biped
  locomotion, in: IEEE/RSJ International Conference on Intelligent Robots and
  Systems (IROS), 2019, pp. 3503--3510.

\bibitem{sim2real2sim}
P.~Chang, T.~Padif, Sim2real2sim: Bridging the gap between simulation and
  real-world in flexible object manipulation, in: IEEE International Conference
  on Robotic Computing (IRC), IEEE, 2020, pp. 56--62.

\bibitem{visual-DR}
T.~Yamanokuchi, Y.~Kwon, Y.~Tsurumine, E.~Uchibe, J.~Morimoto, T.~Matsubara,
  Randomized-to-canonical model predictive control for real-world visual
  robotic manipulation, IEEE Robotics and Automation Letters (RA-L) 7~(4)
  (2022) 8964--8971.

\bibitem{DiDoR}
J.~Brosseit, B.~Hahner, F.~Muratore, M.~Gienger, J.~Peters, Distilled domain
  randomization, arXiv preprint arXiv:2112.03149.

\bibitem{simopt-origin}
Y.~Chebotar, A.~Handa, V.~Makoviychuk, M.~Macklin, J.~Issac, N.~Ratliff,
  D.~Fox, Closing the sim-to-real loop: Adapting simulation randomization with
  real world experience, in: IEEE International Conference on Robotics and
  Automation (ICRA), 2019, pp. 8973--8979.

\bibitem{simopt-baysian}
F.~Muratore, C.~Eilers, M.~Gienger, J.~Peters, Data-efficient domain
  randomization with bayesian optimization, IEEE Robotics and Automation
  Letters (RA-L) 6~(2) (2021) 911--918.

\bibitem{bayessim}
F.~Ramos, R.~C. Possas, D.~Fox, Bayessim: adaptive domain randomization via
  probabilistic inference for robotics simulators, in: Robotics: Science and
  Systems (RSS), 2019.

\bibitem{ADR}
I.~Akkaya, M.~Andrychowicz, M.~Chociej, M.~Litwin, B.~McGrew, A.~Petron,
  A.~Paino, M.~Plappert, G.~Powell, R.~Ribas, et~al., Solving rubik's cube with
  a robot hand, arXiv preprint arXiv:1910.07113.

\bibitem{DnC}
D.~Ghosh, A.~Singh, A.~Rajeswaran, V.~Kumar, S.~Levine, Divide-and-conquer
  reinforcement learning, in: International Conference on Learning
  Representations (ICLR), 2018.

\bibitem{gym}
G.~Brockman, V.~Cheung, L.~Pettersson, J.~Schneider, J.~Schulman, J.~Tang,
  W.~Zaremba, Openai gym, arXiv preprint arXiv:1606.01540.

\bibitem{mujoco}
E.~Todorov, T.~Erez, Y.~Tassa, Mujoco: A physics engine for model-based
  control, in: IEEE/RSJ International Conference on Intelligent Robots and
  Systems (IROS), 2012, pp. 5026--5033.

\bibitem{powder-sim2real}
C.~Matl, Y.~Narang, R.~Bajcsy, F.~Ramos, D.~Fox, Inferring the material
  properties of granular media for robotic tasks, in: IEEE International
  Conference on Robotics and Automation (ICRA), 2020, pp. 2770--2777.

\bibitem{powder-manipulation}
S.~Clarke, T.~Rhodes, C.~G. Atkeson, O.~Kroemer, Learning audio feedback for
  estimating amount and flow of granular material, Proceedings of Machine
  Learning Research (PMLR) 87.

\bibitem{powder-manipulation-learning}
N.~Tuomainen, D.~Blanco-Mulero, V.~Kyrki, Manipulation of granular materials by
  learning particle interactions, IEEE Robotics and Automation Letters (RA-L)
  7~(2) (2022) 5663--5670.

\bibitem{excavator-sim2real}
D.~Jud, P.~Leemann, S.~Kerscher, M.~Hutter, Autonomous free-form trenching
  using a walking excavator, IEEE Robotics and Automation Letters (RA-L) 4~(4)
  (2019) 3208--3215.

\bibitem{excavator-sim2real-RL}
P.~Egli, D.~Gaschen, S.~Kerscher, D.~Jud, M.~Hutter, Soil-adaptive excavation
  using reinforcement learning, IEEE Robotics and Automation Letters (RA-L)
  7~(4) (2022) 9778--9785.

\bibitem{excavator-sim2real-geometric}
Q.~Lu, Y.~Zhu, L.~Zhang, Excavation reinforcement learning using geometric
  representation, IEEE Robotics and Automation Letters (RA-L) 7~(2) (2022)
  4472--4479.

\bibitem{DiffTaichi}
Y.~Hu, L.~Anderson, T.-M. Li, Q.~Sun, N.~Carr, J.~Ragan-Kelley, F.~Durand,
  Difftaichi: Differentiable programming for physical simulation, in:
  International Conference on Learning Representations (ICLR), 2020.

\bibitem{curriculum-all}
X.~Wang, Y.~Chen, W.~Zhu, A survey on curriculum learning, IEEE Transactions on
  Pattern Analysis and Machine Intelligence (TPAMI) 44~(9) (2022) 4555--4576.

\bibitem{curriculum-water}
T.~Zhang, R.~Wang, S.~Wang, Y.~Wang, L.~Cheng, G.~Zheng, Autonomous skill
  learning of water polo ball heading for a robotic fish: Curriculum and
  verification, IEEE Transactions on Cognitive and Developmental Systems (TCDS)
  (2022) 1--1.

\bibitem{CPI}
S.~Kakade, J.~Langford, Approximately optimal approximate reinforcement
  learning, in: International Conference on Machine Learning (ICML), 2002, pp.
  267--274.

\bibitem{LPI}
Y.~Abbasi-Yadkori, P.~L. Bartlett, S.~J. Wright, A fast and reliable policy
  improvement algorithm, in: International Conference on Artificial
  Intelligence and Statistics (AISTATS), 2016, pp. 1338--1346.

\bibitem{SPI}
M.~Pirotta, M.~Restelli, A.~Pecorino, D.~Calandriello, Safe policy iteration,
  in: International Conference on Machine Learning (ICML), 2013, pp. 307--315.

\bibitem{policy-distillation}
A.~A. Rusu, S.~G. Colmenarejo, {\c{C}}.~G{\"u}l{\c{c}}ehre, G.~Desjardins,
  J.~Kirkpatrick, R.~Pascanu, V.~Mnih, K.~Kavukcuoglu, R.~Hadsell, Policy
  distillation, in: International Conference on Learning Representations
  (ICLR), 2016.

\bibitem{PD-compress-evaluate}
S.~Stanton, P.~Izmailov, P.~Kirichenko, A.~A. Alemi, A.~G. Wilson, Does
  knowledge distillation really work?, Neural Information Processing Systems
  (NeurIPS) 34 (2021) 6906--6919.

\bibitem{Distral}
Y.~W. Teh, V.~Bapst, W.~M. Czarnecki, J.~Quan, J.~Kirkpatrick, R.~Hadsell,
  N.~Heess, R.~Pascanu, Distral: Robust multitask reinforcement learning, in:
  Neural Information Processing Systems (NeurIPS), 2017, pp. 4499--4509.

\bibitem{DistAfterLearn}
R.~T. Kalifou, H.~Caselles-Dupr{\'e}, T.~Lesort, T.~Sun, N.~Diaz-Rodriguez,
  D.~Filliat, Continual reinforcement learning deployed in real-life using
  policy distillation and sim2real transfer, in: International Conference on
  Machine Learning (ICML), 2019, pp. 4497--4507.

\bibitem{mult-task-PD}
S.~Omidshafiei, J.~Pazis, C.~Amato, J.~P. How, J.~Vian, Deep decentralized
  multi-task multi-agent reinforcement learning under partial observability,
  in: International Conference on Machine Learning (ICML), 2017, pp.
  2681--2690.

\bibitem{DCPI}
N.~Vieillard, O.~Pietquin, M.~Geist, Deep conservative policy iteration, in:
  Association for the Advancement of Artificial Intelligence (AAAI), 2020, pp.
  6070--6077.

\bibitem{CAC}
L.~Zhu, T.~Kitamura, M.~Takamitsu, Cautious actor-critic, in: Asian Conference
  on Machine Learning (ACML), 2021, pp. 220--235.

\bibitem{DistillingPD}
W.~M. Czarnecki, R.~Pascanu, S.~Osindero, S.~Jayakumar, G.~Swirszcz,
  M.~Jaderberg, Distilling policy distillation, in: International Conference on
  Artificial Intelligence and Statistics (AISTATS), 2019, pp. 1331--1340.

\bibitem{SAC}
T.~Haarnoja, A.~Zhou, P.~Abbeel, S.~Levine, Soft actor-critic: Off-policy
  maximum entropy deep reinforcement learning with a stochastic actor, in:
  International Conference on Machine Learning (ICML), 2018, pp. 1861--1870.

\bibitem{DR-DRL-LSTM}
X.~B. Peng, M.~Andrychowicz, W.~Zaremba, P.~Abbeel, Sim-to-real transfer of
  robotic control with dynamics randomization, in: IEEE International
  Conference on Robotics and Automation (ICRA), 2018, pp. 3803--3810.

\bibitem{Gorila}
A.~Nair, P.~Srinivasan, S.~Blackwell, C.~Alcicek, R.~Fearon, A.~De~Maria,
  V.~Panneershelvam, M.~Suleyman, C.~Beattie, S.~Petersen, et~al., Massively
  parallel methods for deep reinforcement learning, arXiv preprint
  arXiv:1507.04296.

\bibitem{A3C}
V.~Mnih, A.~P. Badia, M.~Mirza, A.~Graves, T.~Lillicrap, T.~Harley, D.~Silver,
  K.~Kavukcuoglu, Asynchronous methods for deep reinforcement learning, in:
  International Conference on Machine Learning (ICML), 2016, pp. 1928--1937.

\bibitem{Ape-X}
D.~Horgan, J.~Quan, D.~Budden, G.~Barth-Maron, M.~Hessel, H.~van Hasselt,
  D.~Silver, Distributed prioritized experience replay, in: International
  Conference on Learning Representations (ICLR), 2018.

\bibitem{TD3}
S.~Fujimoto, H.~Hoof, D.~Meger, Addressing function approximation error in
  actor-critic methods, in: International Conference on Machine Learning
  (ICML), 2018, pp. 1587--1596.

\bibitem{DDPG}
T.~P. Lillicrap, J.~J. Hunt, A.~Pritzel, N.~Heess, T.~Erez, Y.~Tassa,
  D.~Silver, D.~Wierstra, Continuous control with deep reinforcement learning,
  in: International Conference on Learning Representations (ICLR), 2016.

\end{thebibliography}

\end{document}